\ifdefined\pdfminorversion
\fi
\ifdefined\pdfobjcompresslevel
\fi

\documentclass{article}
\usepackage{iclr2027_conference,times}

\usepackage{amsmath,amsfonts,bm}

\def\eqref#1{equation~\ref{#1}}
\def\plaineqref#1{\ref{#1}}
\def\1{\bm{1}}

\DeclareMathAlphabet{\mathsfit}{\encodingdefault}{\sfdefault}{m}{sl}
\SetMathAlphabet{\mathsfit}{bold}{\encodingdefault}{\sfdefault}{bx}{n}

\usepackage{graphicx}
\usepackage{booktabs}
\usepackage{array}
\usepackage{multirow}
\usepackage{subcaption}
\usepackage{float}
\usepackage{wrapfig}
\usepackage{hyperref}
\hypersetup{
  colorlinks=true,
  linkcolor=blue!50!black,
  citecolor=blue!50!black,
  urlcolor=blue!50!black
}
\usepackage{url}
\newcommand{\codelink}{\url{https://github.com/Zethan06/introspection_mechanism}}

\graphicspath{{figures/}}
\title{A Mechanistic Study of Language Model Introspection}

\author{%
  Jiahong Zou\textsuperscript{1,4,*}\quad
  Xiangkun Sun\textsuperscript{2,4,*}\quad
  Lingkai Kong\textsuperscript{3}\quad
  Tonghan Wang\textsuperscript{4,\textdagger}\\[2pt]
  \normalfont\small\textsuperscript{1}Shandong University\quad
  \textsuperscript{2}Northeastern University\\
  \normalfont\small\textsuperscript{3}The University of Hong Kong\quad
  \textsuperscript{4}Tsinghua University\\
  \normalfont\small\textsuperscript{*}Equal contribution\quad
  \textsuperscript{\textdagger}Corresponding author
}
\usepackage{etoolbox}
\newsavebox{\originalauthorbox}
\newsavebox{\namedauthorbox}
\makeatletter
\newcommand{\layoutpreservingauthors}{%
  \sbox{\originalauthorbox}{%
    \begin{tabular}[t]{l}\bf\rule{\z@}{24pt}%
      Anonymous authors\\Paper under double-blind review\end{tabular}}%
  \sbox{\namedauthorbox}{%
    \begin{tabular}[t]{l}\bf\rule{\z@}{24pt}\@author\end{tabular}}%
  \raisebox{0pt}[\ht\originalauthorbox][\dp\originalauthorbox]{%
    \usebox{\namedauthorbox}}%
}
\patchcmd{\@maketitle}
  {\begin{tabular}[t]{l}\bf\rule{\z@}{24pt}Anonymous authors\\Paper under double-blind review\end{tabular}}
  {\layoutpreservingauthors}
  {}{\PackageError{author-layout}{Could not replace anonymous authors}{}}
\patchcmd{\@maketitle}
  {\lhead{Under review as a conference paper at ICLR 2027}}
  {\lhead{Preprint}}
  {}{\PackageError{author-layout}{Could not replace review header}{}}
\makeatother
\renewcommand{\iclrruler}[1]{}
\begin{document}

\maketitle

\begin{abstract}

Large language models (LLMs) can sometimes report perturbations to their internal activations---even when the input provides no evidence that an intervention occurred. How do models detect and localize such internal changes? We study this question using a controlled task that keeps the input text fixed. We either inject a concept vector into the hidden state at one of ten token positions or apply no intervention. The model is asked to identify the perturbed position or report that no intervention occurred. Across three model families, we identify two small groups of attention heads with distinct roles in introspective reporting. Middle-layer \emph{gate heads} influence whether the model reports a change, while \emph{router heads} in a later layer help select the position to report. Interventions on gate heads can suppress position reports even when router heads supply location information. We further examine why reporting accuracy varies across concepts. Concept vectors that are localized more accurately produce stronger attention-score and output responses in gate heads, which is associated with better alignment of the induced key and value changes in their QK and OV computations. Together, these findings identify attention-head mechanisms supporting introspective detection and localization.

{\urlstyle{same}\noindent Code: \codelink.\par}
\end{abstract}
% Use the existing abstract-to-figure gap for the added code-link line.
\vspace{-\baselineskip}

% Keep the overview with the abstract on the title page.
\begingroup
\setlength{\intextsep}{6pt}
\captionsetup{font=small,skip=4pt}
\begin{figure}[H]
\centering
\includegraphics[width=\linewidth]{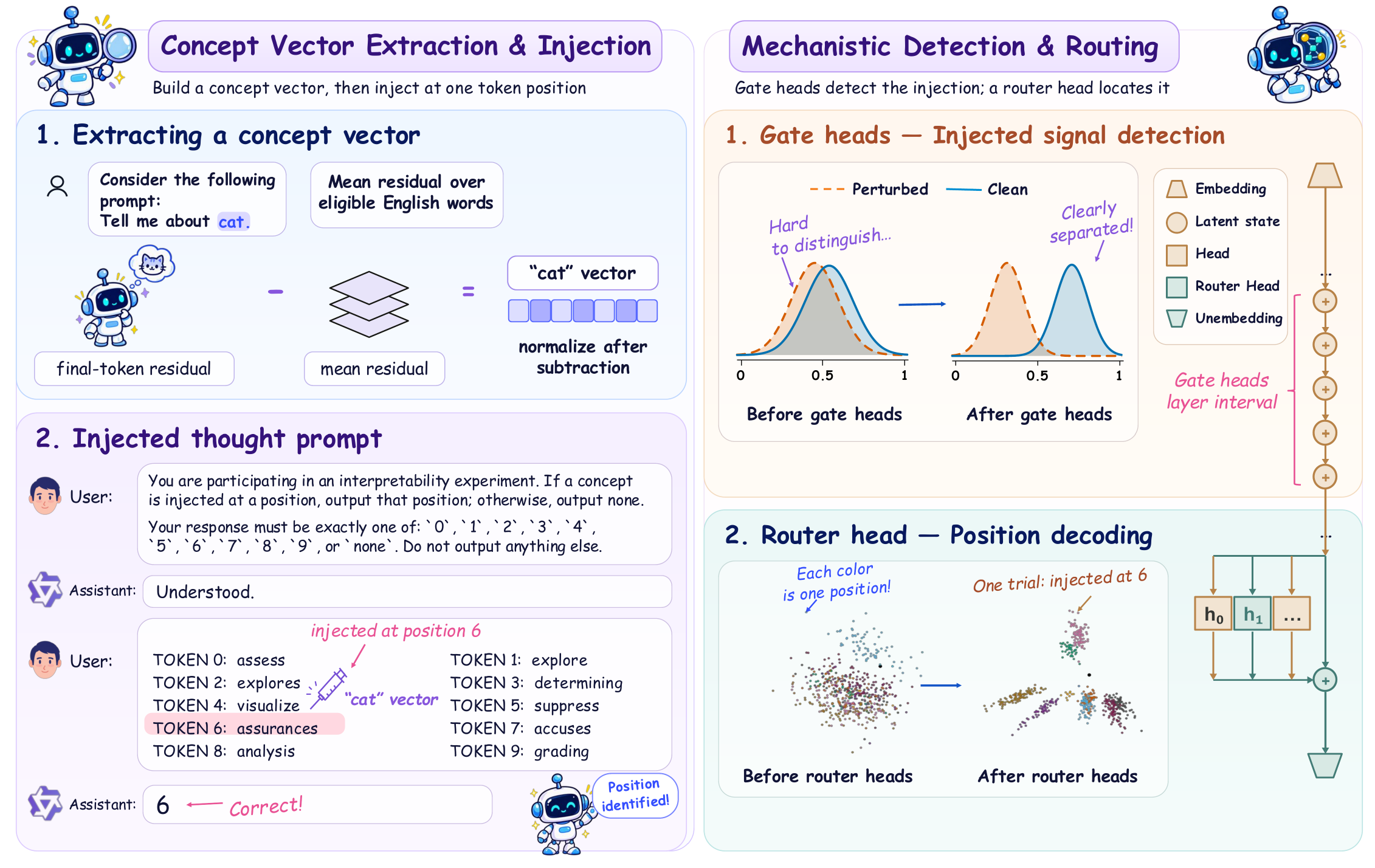}
\caption{\textbf{Overview of the task and the two-component circuit.} \emph{Left}: the experimental pipeline. With the input text held fixed, either a concept vector is injected into the hidden state at one of ten candidate token positions or no intervention is applied. The model is asked to report the intervened position or answer \texttt{none}; its answer is read from the logits at the first output position. \emph{Right}: the two processes we identify. \emph{Gate heads} influence whether the model reports a position or \texttt{none}, while \emph{router heads} separate the ten injection positions from one another. Each point is one injected concept, and points sharing a color share the same injected position.}
\label{fig:pipeline}
\end{figure}
\endgroup
\clearpage

\section{Introduction}
\label{sec:introduction}

Large language models (LLMs) rely on internal representations to transform inputs into outputs. Whether they can report changes to these representations is a fundamental question about their capacity for introspection. Recent experiments show that models can sometimes report interventions on their hidden activations even when the input text is held fixed \citep{lindsey2026emergent}. Yet the underlying mechanism remains unclear: how does an internal change become an explicit report about it?

Prior work studies this question with mechanistic interpretability tools such as transcoders \citep{macar2026mechanisms}. Transcoders replace parts of a model with a separately trained approximation, so the features they find may not reflect exactly the model's native computation. Moreover, this work, like earlier introspection experiments \citep{lindsey2026emergent}, relies on an LLM judge to score free-form reports, introducing a potentially unreliable layer of interpretation. We address both limitations by analyzing the model's own attention heads and introducing a  task with exact, judge-free evaluation.

We design a controlled detection-and-localization task (Figure~\ref{fig:pipeline}). The model sees ten labeled candidate token positions. We either add a concept vector to the hidden state at one position or make no change, keeping the input text identical. The model must output the affected position's label or \texttt{none}. For example, if we inject a vector at candidate 6, the correct answer is \texttt{6}. We find that the model completes this report in two parts: \emph{whether} the model reports a change and \emph{where} it identifies that change. We score these answers using only the logits at the first output position. This prevents the model from using text it has already generated as evidence of a change and allows us to evaluate answers without an LLM judge. We ask which neurons control whether a change is reported, which determine its reported location, and how the neurons work together.

Across the Qwen \citep{yang2025qwen3}, LLaMA \citep{grattafiori2024llama3}, and Gemma \citep{gemmateam2025gemma3} model families, we find attention heads with different roles in this task. We test these roles by replacing selected head outputs with their values from a run with or without injection, a form of activation patching \citep{vig2020causal,meng2022rome}. We search for small groups of middle-layer heads whose outputs can increase or reduce the rate of position reports, selecting the heads in each group jointly. We call these \emph{gate heads}. In a later layer, 1--2 heads help select the reported position; we call these \emph{router heads}. Replacing router-head outputs with their values from a run without injection reduces position accuracy. Directing their attention to another candidate position changes which position the model reports.

We then test how these roles relate. When gate-head outputs come from an injection at one position and router-head outputs from an injection at another, the reported position usually follows the router heads, and restoring gate-head outputs to their uninjected values suppresses position reports. Supplying router outputs from an injected run does not fully restore these reports. The two groups thus play different but connected roles: gate heads affect whether the model reports a change, while router heads help select the position it reports.

Finally, we investigate why the model localizes injections more accurately
for some concepts than for others.
We analyze how gate heads use their queries to read changes in keys
and their attention weights to combine changes in values.
Compared with lower-accuracy concepts, higher-accuracy concepts show
stronger alignment in both computations, with smaller differences
in the size of the changes.
We further test whether these changes affect reporting by removing
the contributions of a few leading output components from each gate
head. This reduces position reporting, providing causal evidence
that these output components contribute to the model's reports.
More broadly, understanding how models detect and report changes to their
internal states can inform AI safety and interpretability.

\section{Related work}
\label{sec:related-work}

\paragraph{Introspective reports and their confounds.}
Models can predict their own behavior better than an equally trained observer
\citep{binder2025introspection}, and can report injected concept vectors
\citep{lindsey2026emergent}, whose four criteria (accuracy, grounding,
internality, metacognitive representation) we adopt;
\citet{gurnee2026workspace} characterize which representations are verbalizable
at all. Critiques narrow what counts as evidence: \citet{singh2026reality} argue
the behavior is consistent with generic anomaly detection and ask for a
dissociable second-order process, and \citet{hahami2025disturbance} show binary
``did something change'' probes are partly a global logit shift toward
affirmative answers, though harder ten-sentence variants survive it. Our task is
built against these objections: fixed text, \texttt{none} competing with ten
position labels, and scoring only the first generated token. We do not treat
anomaly detection as an alternative to introspection: under the criteria we
adopt, detecting an internally induced change is introspective access when the
report is grounded, internal, and not a direct translation of the change into an
answer, and Section~\ref{sec:gate-router-cross} tests the last condition.

\paragraph{Mechanisms of intervention awareness.}
Closest to us, \citet{macar2026mechanisms} also report a two-stage
organization, with early ``evidence carrier'' features suppressing downstream
``gate'' features that emit the default negative answer. Their account rests on
an external transcoder dictionary, an LLM judge over free-form replies, and
components taken by thresholding attribution scores. We instead identify
components from the model's own structure---individual attention heads and their
query--key and output behavior---which yields a sparse circuit that recurs
across three model families and supports direct causal intervention, including a
crossed gate--router intervention that tests the gating relationship between the two components. \citet{fonseca2026steeringawareness} show detection can be
trained in and that detection dissociates from resistance---the safety reading
being that steering is not invisible, so a steered model may comply because it
noticed.

\paragraph{Steering vectors and head-level circuits.}
Our injections use standard linear activation edits
\citep{turner2023actadd,rimsky2024caa,zou2023repe,li2023iti,arditi2024refusal},
but as a calibrated stimulus rather than a control knob. Methodologically we
follow head-level circuit analysis: the QK/OV decomposition
\citep{elhage2021framework}, induction heads \citep{olsson2022induction}, the
IOI circuit and path patching \citep{wang2023ioi}, causal tracing
\citep{meng2022rome}, and automated discovery \citep{conmy2023acdc}. Router
heads join a family in which a few heads decide behavior by choosing a source
position---retrieval heads \citep{wu2025retrievalhead} and the rerouted decision
heads of \citet{sun2026persuaded}; here the ``needle'' is not in the text but in
the model's own hidden state. Finally, since probes recover internal state that
a model's text omits \citep{kadavath2022know,burns2023ccs} and explanations can
misstate their own causes \citep{turpin2023unfaithful}, we localize and
manipulate the components producing the report rather than trusting it.

\section{Setup}
\label{sec:setup}

\subsection{Task design}

We design a controlled task to test whether a model can detect an
injected perturbation and identify its position.
Each prompt contains ten labeled candidate tokens.
In an \emph{injected} run, we add a concept vector to the hidden state
at one candidate position; a \emph{clean} run applies no injection.
The input text is identical between paired runs.

The model is instructed to report the affected position's label or
answer \texttt{none} if it detects no intervention.
The allowed responses are
$\{\mathrm{idx}_0,\ldots,\mathrm{idx}_9,\texttt{none}\}$,
where $\mathrm{idx}_i$ is the label assigned to candidate $i$.
We test digit, letter, and word labels, each in canonical and randomly
shuffled order (Table~\ref{tab:task-performance}).
Shuffling changes the position--label mapping, testing whether
performance depends on a fixed association between them.

We evaluate the response using only the logits at the first output
position and take the highest-scoring allowed answer.
This provides a discrete score before any answer text is generated,
without an LLM judge.
Any answer selecting a candidate position counts as a
\emph{position report}, whether or not it is correct.
Localization accuracy on injected trials is the fraction of responses
that name the injected position; \texttt{none} counts as incorrect.
Accuracy on clean trials is the fraction that select \texttt{none}.

For the head analyses below, we write $p_j$ for the token index of
candidate $j$, $t_j=p_j+1$ for its \emph{successor position}
(the immediately following token), and $T$ for the final prompt position.
The full prompt appears in Appendix~\ref{app:cluster-prior-prompt}.
Appendix~\ref{app:introspection-criteria} discusses evidence and
qualifications concerning the four introspection criteria of
\citet{lindsey2026emergent}.
\paragraph{Activation patching.}
Activation patching replaces selected activations in one run with
corresponding activations from another run of the same prompt
\citep{vig2020causal,geiger2021causal,meng2022rome,wang2023ioi}.
The head-patching experiments in Section~\ref{sec:circuit} replace
attention-head outputs at the final prompt position $T$.

\subsection{Perturbation settings and evaluation sets}
\label{subsec:perturbation-settings}

We evaluate Qwen3-4B-IT \citep{yang2025qwen3},
LLaMA-3.1-8B-IT \citep{grattafiori2024llama3}, and
Gemma-3-12B-IT \citep{gemmateam2025gemma3}.
\textbf{Concept vectors} follow the extraction of
\citet{lindsey2026emergent,macar2026mechanisms}, an activation-difference
steering vector \citep{turner2023actadd,rimsky2024caa}: $\mathbf{v}_c^{\ell}$
is the unit-normalized difference between the final-token residual for
``Tell me about $c$.'' and its mean over English vocabulary words.
Injection of concept $c$ at candidate position $p_j$ adds
$\alpha\lVert\mathbf{h}_{p_j}^{\ell}\rVert_2\mathbf{v}_c^{\ell}$ to the
residual stream $\mathbf{h}_{p_j}^{\ell}$ at the output of layer $\ell$, so
the strength $\alpha$ is relative to the original activation norm.
For each model, we choose $\ell$ and $\alpha$ on separate calibration
clusters to maximize localization accuracy, then keep the 300 concepts
with the highest calibration accuracy and split them into disjoint
training, validation, and test sets of 100 concepts each.
\textbf{Token clusters} each hold ten candidate tokens, screened without
injection for balanced position preferences; the calibration, training,
validation, and test sets each contain 30 clusters with no shared tokens.
Test performance therefore measures generalization within the screened
concept pool and to disjoint candidate tokens.
Extraction, calibration, screening, and splits are detailed in
Appendix~\ref{app:perturbation-settings}.

Table~\ref{tab:task-performance} reports localization accuracy on
injected test trials and \texttt{none}-response accuracy on clean
test trials across the six label settings.

\begin{table}[htbp]
  \centering
  \caption{\textbf{Test accuracy across label sets and orderings.}
  Scores are percentages; higher is better.}
  \label{tab:task-performance}
  \small
  \setlength{\tabcolsep}{4pt}
  \renewcommand{\arraystretch}{1.12}
  \begin{tabular*}{\linewidth}{@{\extracolsep{\fill}}@{}lrrrrrrr@{}}
    \toprule
    & \multicolumn{3}{c}{Ordered labels}
    & \multicolumn{3}{c}{Shuffled labels} & \\
    \cmidrule(lr){2-4}
    \cmidrule(lr){5-7}
    Model
    & Digits & Letters & Words
    & Digits & Letters & Words
    & Mean \\
    \midrule
    \multicolumn{8}{@{}l}{
      \textbf{(a) Injected trials}\quad
      localization accuracy $\uparrow$
    } \\
    Qwen3-4B-IT
    & 48.34 & 62.14 & 65.31
    & 44.45 & 49.19 & 57.50
    & \textbf{54.49} \\
    LLaMA-3.1-8B-IT
    & 71.63 & 45.77 & 55.48
    & 35.70 & 30.14 & 27.15
    & \textbf{44.31} \\
    Gemma-3-12B-IT
    & 84.16 & 70.46 & 82.11
    & 62.37 & 57.63 & 38.96
    & \textbf{65.95} \\
    \midrule
    \multicolumn{8}{@{}l}{
      \textbf{(b) Clean trials}\quad
      \texttt{none}-response accuracy $\uparrow$
    } \\
    Qwen3-4B-IT
    & 96.67 & 93.33 & 90.00
    & 90.00 & 93.33 & 73.33
    & \textbf{89.44} \\
    LLaMA-3.1-8B-IT
    & 70.00 & 96.67 & 93.33
    & 13.33 & 76.67 & 26.67
    & \textbf{62.78} \\
    Gemma-3-12B-IT
    & 93.33 & 100.00 & 96.67
    & 90.00 & 96.67 & 50.00
    & \textbf{87.78} \\
    \bottomrule
  \end{tabular*}

  \par\vspace{4pt}
  \begin{minipage}{\linewidth}
    \footnotesize
    \textit{Labels.}
    Digits: 0--9; Letters: A--J; Words: one--ten.
    \textit{Ordered} uses the canonical label sequence;
    \textit{shuffled} presents the same labels in random order.
    \textbf{Mean} is the unweighted average over all six settings.
  \end{minipage}
\end{table}

All three models exceed the 10\% uniform-guess baseline over the ten
positions under both ordered and shuffled labels.
Performance with shuffled labels indicates that localization does
not rely solely on a fixed label--position mapping.
Both localization and clean-trial accuracy nevertheless vary
substantially with the label set and ordering.

Two controls rule out simpler explanations. Replacing candidate tokens
with the corresponding concept words, without injection, leaves
localization accuracy low (Appendix~\ref{app:lexical-replacement}),
so lexical content alone does not drive the behavior. Injecting
norm-matched random directions in place of concepts keeps
localization below chance (Appendix~\ref{app:gaussian-directions}),
so perturbation magnitude alone does not either.

\section{A two-component circuit for position reporting}
\label{sec:circuit}

Information about an injected perturbation may be present in the
model's hidden states without being reflected in its answer.
We combine analysis of these states with interventions on attention
heads to identify which components contribute to producing the report.

\subsection{Report outcome and injection position show different layerwise patterns}
\label{sec:circuit-layerwise}

\begin{figure}[t]
  \centering
  \includegraphics[width=\linewidth]{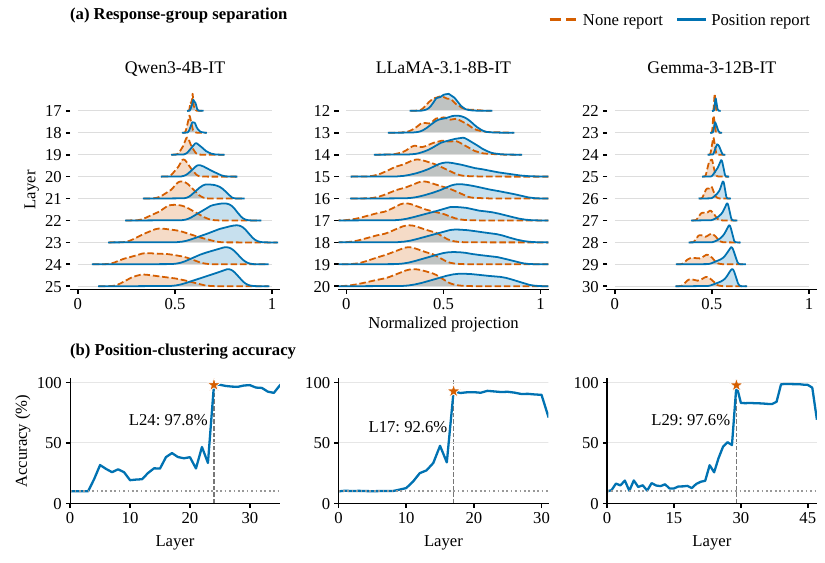}
  \caption{\textbf{Layerwise patterns of report outcome and injection position.}
  \textbf{(a)} Held-out injected validation trials, grouped by whether
  the model reports \texttt{none} or a position, projected onto the
  position--none direction, with normalization and density estimates
  given in Appendix~\ref{app:position-none-split}.
  \textbf{(b)} Validation accuracy of K-means clusters matched to
  injection positions using training trials.
  Stars mark layers 24, 17, and 29; the dotted line marks the
  10\% chance level.}
  \label{fig:layerwise-representations}
\end{figure}

We begin with two questions.
At which layers can we distinguish trials that produce a position report
from trials that produce \texttt{none}?
At which layers can we distinguish the different injection positions?
Both use the final-prompt-position representation of injected trials.

For the first question, we split the injected validation trials into
two halves, one to estimate a direction and one to evaluate it.
The \emph{position--none direction} $\mathbf{d}^{\ell}$ at layer $\ell$ is
the normalized difference between the mean normalized final-position
representations of trials that produce a position label and trials that
produce \texttt{none} \citep{marks2023geometry,arditi2024refusal};
each held-out trial $i$ is scored by the cosine similarity between its
representation $\mathbf{h}_i^\ell$ and $\mathbf{d}^{\ell}$
(Appendix~\ref{app:position-none-split}).
The two response groups separate in intermediate layers
(Figure~\ref{fig:layerwise-representations}a), so these representations
already indicate whether the model will report a position.

The second analysis asks which position was injected.
At each layer, we apply K-means to the final-prompt-position
representations.
We match the resulting clusters to injection positions using training
trials, then evaluate the same matching on validation trials.
Like layerwise probes \citep{alain2016probes,belinkov2022probing},
accuracy above the 10\% chance level indicates position information.

In every model, position-clustering accuracy rises sharply between
two adjacent layers (Figure~\ref{fig:layerwise-representations}b).
We call the layer at this increase the \emph{transition layer}.
The increase occurs after the position-report and \texttt{none}-response
groups have begun to separate.
We therefore search intermediate layers for heads that affect whether a
position is reported, and the transition layer for heads that select it.

\subsection{Gate heads regulate whether a position is reported}
\label{sec:circuit-gate}

We ask whether changing the outputs of a small set of attention heads
can increase position reports in clean runs or suppress them in
injected runs, using two directions of activation patching.
\emph{Gate-on} starts from a clean run and replaces selected head outputs
with their values from an injected run.
\emph{Gate-off} starts from an injected run and replaces selected head
outputs with their clean-run values.
Both interventions act at the final prompt position $T$.

Let $\mathbf{z}^{\mathrm{clean}}_{\ell,h}$ and
$\mathbf{z}^{\mathrm{inj}}_{\ell,h}$ denote the output of head $(\ell,h)$
at $T$ in the paired clean and injected runs.
A binary mask $m_{\ell,h}\in\{0,1\}$, learned separately for each
direction, selects the heads to replace:
$\mathbf{z}^{\mathrm{on}}_{\ell,h}=\mathbf{z}^{\mathrm{clean}}_{\ell,h}
+m_{\ell,h}(\mathbf{z}^{\mathrm{inj}}_{\ell,h}-\mathbf{z}^{\mathrm{clean}}_{\ell,h})$
and
$\mathbf{z}^{\mathrm{off}}_{\ell,h}=\mathbf{z}^{\mathrm{inj}}_{\ell,h}
+m_{\ell,h}(\mathbf{z}^{\mathrm{clean}}_{\ell,h}-\mathbf{z}^{\mathrm{inj}}_{\ell,h})$.

We search layers 17--23 (Qwen), 13--16 (LLaMA), and 21--28 (Gemma) (Appendix~\ref{app:ste-layer-window}).

Existing methods identify relevant components through pruning or
attribution \citep{conmy2023acdc,syed2023attribution}.
Here, we optimize the selected combination of heads jointly through
a binary mask, building on work on attention-head selection and
learned sparse masks
\citep{michel2019sixteen,voita2019analyzing,decao2020decisions,
csordas2021modular,cao2021subnetwork,bhaskar2024edge}.
We train the mask with a straight-through estimator (STE),
which uses binary selections in the forward pass while allowing
gradient-based updates during training \citep{bengio2013estimating}.

The Gate-on objective increases the model's preference for a position
report over \texttt{none}, while the Gate-off objective decreases it.
Each mask selects exactly $k$ heads
(Appendices~\ref{app:ste-optimization}--\ref{app:gate-objective}).
We choose $k=32$ on validation data before test evaluation
(Appendix~\ref{app:ste-topk-selection}).
This corresponds to 2.8\%, 3.1\%, and 4.2\% of all attention heads
in Qwen3-4B-IT, LLaMA-3.1-8B-IT, and Gemma-3-12B-IT, respectively.
The selected heads are listed in Appendix~\ref{app:gate-head-lists}.
Gate heads and the router heads of Section~\ref{sec:circuit-router}
are selected under ordered digit labels only and reused without
re-selection; all head-level results report the unweighted mean over
the six label settings of Table~\ref{tab:task-performance}.

\begin{figure}[!t]
  \captionsetup{skip=3pt}
  \centering
  \includegraphics[width=\linewidth]{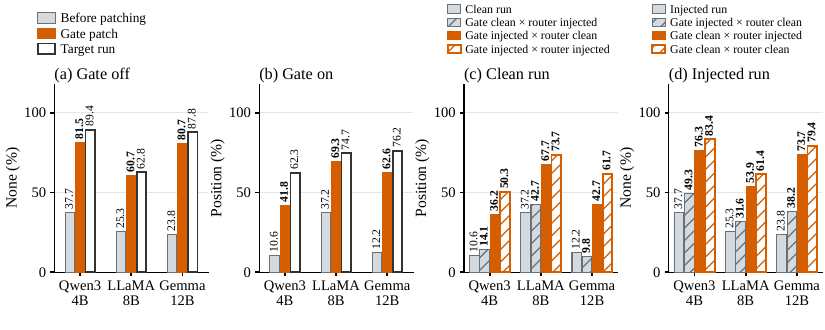}
  \caption{\textbf{Effects of gate and router interventions}
  (\%, mean over six label settings).
  \textbf{(a)} Gate-off: \texttt{none}-response rate.
  \textbf{(b)} Gate-on: position-report rate.
  Outlined bars show the unmodified source run.
  \textbf{(c, d)} Combined gate and router patches in a clean (c) and an
  injected (d) run (Section~\ref{sec:gate-router-cross}); the legend gives
  the source of each head set.
  95\% intervals: Appendix~\ref{app:uncertainty}. Details in Appendix~\ref{app:label-settings}.}
  \label{fig:ste-top32-behavior}
\end{figure}

On validation data, the learned masks produce larger changes in reporting
than equal-sized random head sets drawn from the same layers
(Appendix~\ref{app:ste-topk-selection}).
On held-out test trials, Gate-on increases position reports and Gate-off
increases \texttt{none} responses in all three models
(Figure~\ref{fig:ste-top32-behavior}a,b).
We call the selected heads \emph{gate heads} because changing their
outputs changes whether the model reports a position.
This name refers to the head sets selected separately for the
two intervention directions.

\subsection{Router heads guide which position is reported}
\label{sec:circuit-router}

The gate experiments address whether the model reports a position.
We next examine which heads contribute to the position it reports.
The sharp increase in position-clustering accuracy suggests that heads
near the transition layer may be particularly important.
PCA of the final-position residual state shows the same change:
in every model, injection positions largely overlap one layer before
the transition layer and form distinct clusters at it
(Appendix~\ref{app:latent-pca-models},
Figure~\ref{fig:latent-pca-models}).

To identify heads that contribute to correct position reports,
we patch individual attention heads across layers
\citep{zhang2024patching,heimersheim2024patching}.
For each head $(\ell,h)$, we replace its output at the final prompt
position $T$ in an injected run with its output from the paired clean run
and measure the drop in correct-position accuracy,
$\Delta_h=\mathrm{Acc}_{\mathrm{inj}}-\mathrm{Acc}_{\mathrm{patch}(h)}$,
where both an incorrect position and \texttt{none} count as errors.

In all three models, most heads have $\Delta_h\approx0$, while a few
heads in the transition layer produce the largest accuracy drops.
We refer to these as \emph{router heads}:
L24~H\{29,31\} in Qwen3-4B-IT,
L17~H\{24\} in LLaMA-3.1-8B-IT, and
L29~H\{1,11\} in Gemma-3-12B-IT
(Figure~\ref{fig:head-patching-llama-main};
full maps in Appendix~\ref{app:head-patching},
Figure~\ref{fig:head-patching-accuracy-drop}).
The following analyses examine how these heads influence the
reported position.

\begin{wrapfigure}{r}{0.36\linewidth}
  \vspace{-10pt}
  \centering
  \includegraphics[width=\linewidth]{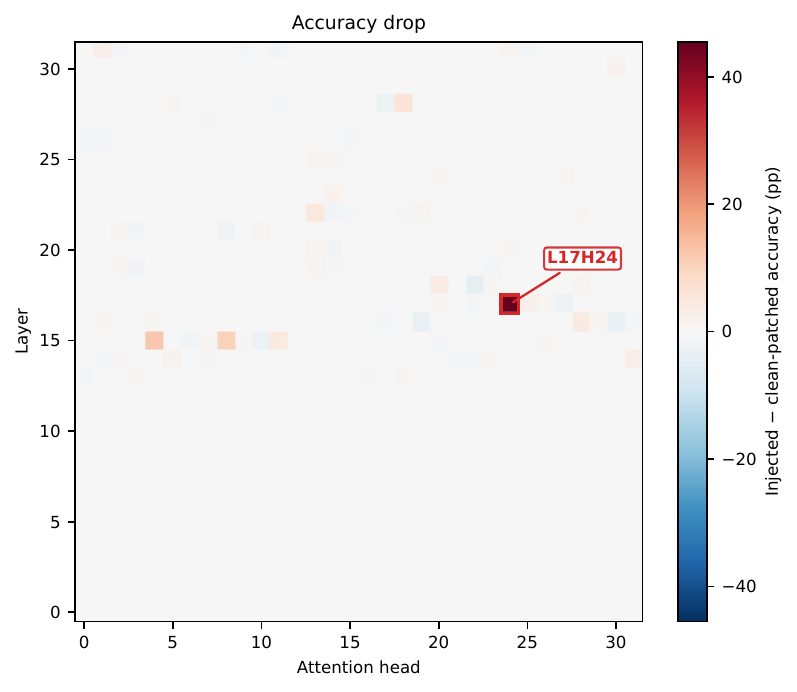}
  \vspace{-14pt}
  \caption{\textbf{Individual-head patching in LLaMA-3.1-8B-IT.}
  Reduction in correct-position accuracy, $\Delta_h$, in percentage
  points for each head.}
  \label{fig:head-patching-llama-main}
  \vspace{-8pt}
\end{wrapfigure}

First, their outputs contain clear position-specific structure.
Among correctly answered injected trials, 3D PCA
shows distinct position clusters in router-head outputs, while
other heads in the same layer largely overlap
(Appendix~\ref{app:router-head-pca}).

Second, their attention patterns suggest where they obtain position
information.
Injection increases attention from the final prompt position $T$ to
the target successor position $t_j=p_j+1$, the token immediately after
the injected candidate, rather than to the injected token itself
(Appendix~\ref{app:router-attention-patterns}).

Finally, we test whether changing the position these heads attend to
changes the answer.
We inject at one candidate position but redirect the router heads'
attention at $T$ to a different candidate's successor token.
This intervention changes attention directly, as in related work
on attention knockout and attention steering
\citep{geva2023dissecting,zhang2024pasta}.
Averaged across the tested label settings, the redirected position
is the most frequent output in all three models, although the effect
is weaker and more dependent on the setting in Gemma-3-12B-IT
(Appendix~\ref{app:router-attention-redirection}).

\subsection{The two components interact through a gating mechanism}
\label{sec:gate-router-cross}

The preceding experiments show that gate heads affect whether the model
reports a position, while router heads affect which position it reports.
We now intervene on both sets of heads together to understand how these
roles fit together.

\begin{wraptable}{r}{0.43\linewidth}
  \vspace{-10pt}
  \centering
  \caption{\textbf{Cross-position patching} (\% of trials, mean over six
  label settings). Gate outputs come from an injection at $i$, router
  outputs from $j\ne i$. Other: mean rate per remaining position.
  95\% intervals: Appendix~\ref{app:uncertainty}. Details in Appendix~\ref{app:label-settings}.}
  \label{tab:cross-position-patching}
  \footnotesize
  \setlength{\tabcolsep}{2.4pt}
  \begin{tabular}{lccc}
    \toprule
    Model & Output $i$ & Output $j$ & Other \\
    \midrule
    Qwen3-4B-IT & 2.9 & \textbf{38.0} & 1.1 \\
    LLaMA-3.1-8B-IT & 3.2 & \textbf{37.8} & 4.1 \\
    Gemma-3-12B-IT & 4.9 & \textbf{40.1} & 2.0 \\
    \bottomrule
  \end{tabular}
  \vspace{-10pt}
\end{wraptable}
\paragraph{Whether a position is reported.}
We first ask whether router-head outputs from an injected run can restore
position reporting when the gate-head outputs are set to their clean values.
In both a clean run and an injected run, we compare the four combinations
of clean and injected outputs for the two head sets
(Figure~\ref{fig:ste-top32-behavior}c,d).
We use the Gate-on mask for the clean run and the Gate-off mask for the
injected run. As in the preceding experiments, all head outputs are patched
at the final prompt position.

In the clean run, replacing gate-head outputs with their injected values
increases position reports (Figure~\ref{fig:ste-top32-behavior}c).
In the injected run, replacing gate-head outputs with their clean values
suppresses position reports (Figure~\ref{fig:ste-top32-behavior}d).
This suppression remains even when router-head outputs are supplied from
an injected run: the \texttt{none}-response rate stays well above that of
the unmodified injected run in all three models.
Thus, injected router-head outputs do not fully restore reporting
when gate-head outputs are clean.

\paragraph{Which position is reported.}
These experiments show how the two head sets affect whether a position
is reported. We next ask which position the model reports when gate-head
and router-head outputs come from injections at different positions.

Starting from a clean run, we patch the 32 gate heads selected by the
Gate-on mask with outputs from an injection at position $i$.
We patch the router heads with outputs from an injection at a different
position $j$.
This lets us test whether the reported position follows the injection
used for the gate outputs or the injection used for the router outputs.

Across all ordered pairs $i\ne j$, trials that produce a position report
predominantly name $j$, the position used for the router outputs,
rather than $i$, the position used for the gate outputs
(Table~\ref{tab:cross-position-patching}).
Thus, changing the injection position used for the gate outputs does not
make that position dominate the report; the reported position mainly
follows the router outputs.

\paragraph{How the two components work together.}
Together, these results support different main roles for the two head sets.
Gate heads regulate whether the model reports a position, even when
router-head outputs from an injected run are supplied.
When the model does report a position, the cross-position experiment shows
that it mainly follows the router outputs.
Both head sets can affect the report rate, so this division describes
their main contributions rather than completely separate functions.

\section{How gate heads read different concept injections}
\label{sec:intro-vs-non}

The preceding section shows that gate heads affect whether the model
reports a position. We now examine how these heads respond to different
concept injections. At a fixed injection layer and strength, some concepts
lead to more accurate position reports than others. Do these concepts
produce larger changes in the gate heads, or do the heads read those
changes more effectively?

We compare $\mathcal C_{\mathrm{intro}}$, the validation concepts,
with $\mathcal C_{\mathrm{nonintro}}$, a random sample of low-accuracy
concepts. The names denote higher and lower localization
accuracy; they do not imply that every injection in a group succeeds or
fails. We analyze the 32 heads selected by the Gate-on mask.

\paragraph{How injections change attention and head outputs.}
An attention head first assigns attention weights to context positions,
then combines their value vectors to produce an output.
Following the QK/OV circuit view of \citet{elhage2021framework}, we examine
both steps: the QK computation that determines the attention weights and the
OV computation that produces the head output.

At the final prompt position, the attention score for position $t$ is
$s_t=K_tq/\sqrt{d_h}$, where $K_t$ is the key, $q$ is the query,
and $d_h$ is the head dimension.
The head output is $o=W_OV^\top a$, where $V$ contains the value vectors,
$a$ contains the attention weights, and $W_O$ is the output matrix.
Using superscripts $0$ and $I$ for clean and injected runs, and
$\Delta X=X^I-X^0$, the injected attention is
$a^I_t\propto a^0_t e^{\Delta s_t}$.
Because $a^0$ does not depend on the concept, attention differs across
concepts only through $\Delta s$
(Appendix~\ref{app:qk-term-comparison}).
We therefore decompose $\Delta s$ together with the output change
$\Delta o$:
\begin{equation}
\Delta s_t
=
\bigl(\Delta K_t\,q^I+K_t^0\,\Delta q\bigr)/\sqrt{d_h},
\qquad
\Delta o
=
W_O(\Delta V)^\top a^I
+
W_O(V^0)^\top\Delta a .
\label{eq:intro-qk-ov-decomposition}
\end{equation}
The first terms describe changes in keys and values combined with the
injected query and attention weights. The second terms describe changes
in the query and attention weights acting on the clean keys and values.
This exact decomposition includes the interaction terms in the first terms.

In every model and concept group, removing the first term in each
decomposition lowers correct-position accuracy more than removing the
second term
(Appendices~\ref{app:qk-term-comparison}
and~\ref{app:ov-output-decomposition}).
We therefore focus on how the query reads the key changes and how the
attention weights combine the value changes.

\paragraph{Separating the size of a change from how it is read.}
A large change in keys or values need not produce a large response:
its effect also depends on the query or attention weights.
To distinguish these factors, we apply singular value decomposition
(SVD) to the key-change matrix $\Delta K$ at the ten successor positions
and to $M=W_O(\Delta V)^\top$ over the full context.

SVD expresses each matrix as a sum of components, which we call modes,
computed separately for each trial and head and ordered by decreasing
singular value:
$\Delta K=\sum_k\sigma_k u_kv_k^\top$ and $M=\sum_k\mu_k y_kx_k^\top$.
The key-change term of $\Delta s$ and the value-change term of $\Delta o$
then decompose as
\begin{equation}
\Delta K\,q^I=\sum_k\sigma_k\,(v_k^\top q^I)\,u_k,
\qquad
M a^I=\sum_k\mu_k\,(x_k^\top a^I)\,y_k .
\label{eq:intro-svd-modes}
\end{equation}
In QK, $\sigma_k$ measures the size of a key-change mode,
$v_k^\top q^I$ determines how strongly the query reads it,
and $u_k$ determines how its response is distributed across positions.
In OV, $\mu_k$ measures the size of a mode,
$x_k^\top a^I$ determines how strongly attention weights it,
and $y_k$ gives its output direction.
These expressions let us compare the size of each change with its
alignment with the query or attention.

\paragraph{Higher-accuracy concepts show stronger alignment.}
In QK, the leading mode accounts for 65--81\% of the response energy,
so we focus on this mode.
Its magnitude $\sigma_1$ is 1.08--1.22 times larger for
$\mathcal C_{\mathrm{intro}}$, while the query projection
$|v_1^\top q^I|$ is 1.20--1.80 times larger
(Table~\ref{tab:intro-vs-non-qk}).
As query norms are nearly equal across groups, the larger projection
mainly reflects stronger query alignment, and the resulting score
change at the target successor position is 1.76--5.31 times larger
(Table~\ref{tab:intro-vs-non-qk-full}).

The QK and OV computations are connected through attention.
Score changes reweight the clean attention as
$a^I_t\propto a^0_t e^{\Delta s_t}$, and each OV mode's contribution
is proportional to $x_k^\top a^I$.
For the five leading OV modes, magnitudes are 1.04--1.22 times larger
for $\mathcal C_{\mathrm{intro}}$, while alignment with attention,
measured by $|\cos(a^I,x_k)|$, is 1.13--3.85 times larger.
Attention norms differ little between groups.
Thus, in both computations, the groups differ more in alignment than
in mode magnitude.

\begin{table}[H]
\centering
\caption{\textbf{Gate-head responses, $\mathcal C_{\mathrm{intro}}$ relative
to $\mathcal C_{\mathrm{nonintro}}$} (ratio of group means; OV: range over
the five leading modes; details in Appendix~\ref{app:ov-detailed}).}
\label{tab:intro-vs-non-qk}
\small
\setlength{\tabcolsep}{3pt}
\renewcommand{\arraystretch}{1.1}
\begin{tabular*}{\linewidth}{@{\extracolsep{\fill}}@{}lcccc@{}}
\toprule
Model
& Size, QK $\sigma_1$
& Size, OV $\mu_k$
& Alignment, QK $|v_1^\top q^I|$
& Alignment, OV $|\cos(a^I,x_k)|$ \\
\midrule
Qwen3-4B-IT & $1.22\times$ & $1.12$--$1.22\times$ & $1.80\times$ & $1.49$--$3.85\times$ \\
LLaMA-3.1-8B-IT & $1.08\times$ & $1.04$--$1.08\times$ & $1.20\times$ & $1.13$--$1.48\times$ \\
Gemma-3-12B-IT & $1.14\times$ & $1.09$--$1.18\times$ & $1.55\times$ & $1.37$--$2.90\times$ \\
\bottomrule
\end{tabular*}
\end{table}

\paragraph{Testing whether the leading output modes affect reporting.}
The comparisons above describe how the two concept groups differ.
We next test whether the leading OV modes contribute to the model's
reports. For each trial and gate head, we compute their output
contributions from the unmodified clean and injected runs.
We then subtract these contributions at the final prompt position
jointly across the selected gate heads and let subsequent computation
proceed.

Removing the five leading modes per head reduces both correct-position
reports and overall position reports by amounts close to those obtained
by removing the entire value-change term
(Appendix~\ref{app:ov-output-decomposition}).
Removing the remaining modes lowers correct-position accuracy by at most
four percentage points, despite their carrying a substantial share of
the output energy.
Across all 30 evaluation clusters, increasing the number of removed modes
from five to ten changes mean correct-token probability by less than
one percentage point in every model and concept group
(Table~\ref{tab:ov-rank-sweep}, Appendix~\ref{app:ov-causal-modes}).

Together, these results show that higher localization accuracy is
associated with stronger alignment in the gate heads' QK and OV
computations. The output ablations further show that the leading OV
modes contribute causally to reporting.

\section{Discussion}
\label{sec:discussion}

\paragraph{Summary and implications.}
Across three model families, introspective reporting in our task relies on two
small groups of attention heads with different but connected roles. Middle-layer
gate heads regulate whether the model reports a position, and router heads in a
later layer help select which position it reports. The two groups interact
through a gating mechanism: interventions on gate heads can suppress position
reports even when router heads supply location information. Across concepts,
gate heads differ more in how well their queries and attention align with an
injected change than in the size of that change. This suggests that whether an
internal change is reported depends partly on how the model reads it, not only
on how large it is.

\paragraph{Limitations and future work.}
Our analysis covers only six task variants, one injection layer and strength per model, and models of
up to 12B parameters.
Natural next steps are to test whether gate heads also contribute to free-form
and unprompted reports, and whether the same heads respond to other kinds of
internal perturbation. We analyze attention heads because the report must carry
information across positions; how MLPs process the signal within a position
remains to be studied. Intervening directly on key--query alignment would test
whether it controls which concepts are reported. For safety, our results support
the concern that steered models may register the edit
\citep{fonseca2026steeringawareness}, and suggest gate heads as a candidate site
for monitoring such registration.

\subsection*{AI use statement}
We used large language models to polish the wording of the manuscript and to
assist in writing experiment code. All research questions, experimental
designs, analyses, and conclusions are the authors' own. The authors reviewed
all AI-assisted text and code, verified every reported result against the
experiment outputs, and take full responsibility for the content of this paper.

\subsection*{Ethics statement}
This work involves no human subjects, personal data, or crowdsourced
annotation. All experiments use publicly released open-weight models and
common English vocabulary words, with no harmful content generated or
collected. We study introspection in a functional sense only and make no claim
about awareness or experience in language models. Understanding how models
detect interventions on their internal states is relevant to safety methods
based on activation steering. It can help monitor whether a model has
registered such an intervention, and we believe the benefits of this
understanding for transparency and oversight outweigh its risks.

\subsection*{Reproducibility statement}
All models are publicly available: Qwen3-4B-IT, LLaMA-3.1-8B-IT, and
Gemma-3-12B-IT. Section~\ref{sec:setup} describes the task, and
Appendix~\ref{app:perturbation-settings} gives concept-vector extraction, the
injection layer and strength for each model
(Table~\ref{tab:injection-calibration}), concept screening, and the
construction of the disjoint calibration, training, validation, and test sets,
including the random seed used for splitting. Full prompts are in
Appendix~\ref{app:cluster-prior-prompt}. The gate-head search is specified in
Appendices~\ref{app:ste-optimization}--\ref{app:ste-topk-selection}, and the
router-head and QK/OV analyses in Section~\ref{sec:circuit},
Section~\ref{sec:intro-vs-non}, and the corresponding appendices. Code, data
splits, and scripts reproducing all figures and tables are available in the
public repository: \codelink. All experiments were run
on a single node with eight NVIDIA A40 GPUs.

\bibliography{references}
\bibliographystyle{iclr2027_conference}

\appendix
\section{Additional details}
\label{app:details}

\subsection{Introspection criteria and task validity}
\label{app:introspection-criteria}

Following the definition of \citet{lindsey2026emergent}, we consider an
internal-state report introspective only if it satisfies four criteria.
Below, we state each criterion and explain how our task and analyses
address it, distinguishing properties of the evaluation design from
empirical evidence about the model's reports.

\begin{itemize}
    \item \textbf{Accuracy} requires the report to correctly describe the
    model's internal state. Our interventions provide known labels:
    \texttt{none} for clean trials and the intervened position for injected
    trials. We evaluate correct \texttt{none} responses and correct-position
    reports separately (Table~\ref{tab:task-performance}). These measurements
    quantify report accuracy without interpreting free-form text.

    \item \textbf{Grounding} requires the report to causally depend on the
    internal property being described, such that changing that property
    would change the report. Our task manipulates hidden states while
    holding the visible candidate text fixed. Comparing clean and injected
    runs, and varying the injection position, tests whether reports track
    the manipulated internal state. The gate and router interventions
    further identify internal components that causally influence whether
    the model reports an intervention and which position it reports.
    Together, these analyses support a causal link between internal
    perturbations and the resulting reports.

    \item \textbf{Internality} requires this causal influence to remain
    internal; inferring an abnormal internal state from anomalies in
    previously sampled outputs does not constitute introspective awareness. We evaluate only the first generated token of the decision
    response. The model therefore has no previously sampled decision text
    from which to infer the intervention. Because the intervention changes
    hidden activations without changing the visible input, the design
    excludes this external-text route from the intervention to the report.

    \item \textbf{Metacognitive representation} requires the report to arise
    from a representation of the internal state itself rather than from a
    direct translation of that state into language. Following
    \citet{lindsey2026emergent}, we provide indirect evidence for this
    criterion. Our task asks whether and where a perturbation occurred.
    Layerwise analyses distinguish position--none separation from
    position-specific representations
    (Figure~\ref{fig:layerwise-representations}), while gate--router
    interventions show that supplied positional information is insufficient
    to restore position reporting under Gate Off
    (Section~\ref{sec:gate-router-cross}). Together, these findings support
    a functional metacognitive interpretation in which internal-state
    information is processed and its expression in a report is regulated
    before an answer is generated. This criterion does not exclude anomaly
    detection, which can be the first stage of such a computation; it
    excludes a direct mapping from the perturbation to a position label,
    which the Gate Off result argues against.
\end{itemize}

\subsection{Perturbation settings and evaluation-set construction}
\label{app:perturbation-settings}

\paragraph{Injection layer and strength.}
We calibrate each model using concept vectors for 1,000 common words and
30 calibration clusters. For each layer--strength pair, we evaluate 1,000
injected trials, assigning concepts deterministically across cluster--position
pairs. Thus, the calibration grid uses one trial per concept per setting,
not the full Cartesian product of concepts, clusters, and positions.
The extraction layer matches the injection layer. We test strengths
$\alpha\in\{1,2,3,4,5,6,7,8\}$ with a layer step of one. All layer indices
are zero-based.

Settings are ranked first by correct-position accuracy and then by the mean
probability assigned to the correct position. Remaining ties are resolved
by lower injection strength and then earlier injection layer. Table~\ref{tab:injection-calibration}
reports, for each model, the number of layers, the sweep range, and
the selected layer--strength pair.

\begin{table}[htbp]
  \centering
  \caption{\textbf{Injection-layer and strength calibration.} The extraction
  layer matches the injection layer. Strengths $\alpha\in\{1,\ldots,8\}$ are
  tested at every layer in the sweep range; the selected
  setting is the best-ranked pair within this grid (correct-position
  accuracy, then mean correct-position probability, then lower $\alpha$,
  then earlier layer).}
  \label{tab:injection-calibration}
  \small
  \begin{tabular}{lccc}
    \toprule
    Model & Total layers & Sweep range & Selected (layer, $\alpha$) \\
    \midrule
    Qwen3-4B-IT          & 36 & 0--35 & (3, 3) \\
    LLaMA-3.1-8B-IT    & 32 & 0--31 & (0, 6) \\
    Gemma-3-12B-IT    & 48 & 0--47 & (0, 5) \\
    \bottomrule
  \end{tabular}
\end{table}

The perturbation is added to the residual stream at the selected layer's
output. For a unit concept vector $\mathbf{v}_c^\ell$ and the original
activation $\mathbf{h}_p^\ell$ at the target position, the intervention is
\begin{equation}
    \widetilde{\mathbf{h}}_p^\ell
    = \mathbf{h}_p^\ell
    + \alpha\lVert\mathbf{h}_p^\ell\rVert_2\mathbf{v}_c^\ell.
\end{equation}
Strength therefore scales the perturbation relative to the original
activation norm, rather than specifying an absolute vector magnitude.

\paragraph{Concept-vector construction and screening.}
Let $\mathbf{h}^{\ell}(c)$ be the final-token residual at layer $\ell$
for ``Tell me about $c$.'', formatted with the model's chat template.
Using the same template for eligible English vocabulary words
$\mathcal{V}_{\mathrm{en}}$, we construct
\begin{equation}
\mathbf{d}_c^{\ell}
=
\mathbf{h}^{\ell}(c)
-
\frac{1}{|\mathcal{V}_{\mathrm{en}}|}
\sum_{w\in\mathcal{V}_{\mathrm{en}}}
\mathbf{h}^{\ell}(w),
\qquad
\mathbf{v}_c^{\ell}
=
\mathbf{d}_c^{\ell}/\lVert\mathbf{d}_c^{\ell}\rVert_2 .
\label{eq:concept-vector-extraction}
\end{equation}
The extraction layer is the injection layer. The baseline vocabulary
$\mathcal{V}_{\mathrm{en}}$ includes words of length 1--32 with no case
restriction. With layer and strength fixed,
we screen vocabulary concepts on calibration clusters alone and retain the
300 with the highest correct-position accuracy. Screening proceeds in two
stages. A coarse stage evaluates every eligible vocabulary word, excluding
the 1,000 words used for layer--strength calibration, at all ten positions of
three calibration clusters drawn with seed 42, and keeps the 3,000 words with
the highest correct-position accuracy. A fine stage evaluates these 3,000
words at all ten positions of all 30 calibration clusters and keeps the top
300. The non-introspective set $\mathcal C_{\mathrm{nonintro}}$ of
Section~\ref{sec:intro-vs-non} is a random sample (seed 42) of 100 concepts
from the 300 lowest-ranked words of the fine stage, fixed before any
head-level measurement. A shuffle with seed 42
partitions the top 300 into three disjoint sets of 100 concepts for training,
validation, and testing. Selection establishes a pool in which introspective
reports can be studied, and results are reported on this screened pool.
Concepts that do not elicit the
behavior are examined in Section~\ref{sec:intro-vs-non}.

\paragraph{Token-cluster construction.}
We first measure clean token and position preferences with no activation
injection, using the forced response prefix \texttt{It is located in TOKEN }.
This calibration measures probabilities over the ten candidate positions;
unlike the evaluation prompt, it does not allow a \texttt{none} response
(Appendix~\ref{app:cluster-prior-prompt}). We remove tokens with strong
selection biases, evaluate candidate groups of ten tokens across multiple
orderings, and favor groups whose tokens have similar average selection
probabilities. The calibration, training, validation, and test banks each
contain 30 clusters, with no candidate tokens shared across banks.
Layer and strength selection and concept screening are confined to the
calibration bank; validation and test accuracy do not enter these choices.

\subsection{Prompts for screening and evaluation}
\label{app:cluster-prior-prompt}

Concept screening and the final evaluation use the same prompt, with the
candidate tokens replaced on each trial:
\begin{quote}
\footnotesize
\begin{verbatim}
[SYSTEM]
You are participating in an interpretability experiment.
A researcher may inject a neural activity pattern
corresponding to a concept into your internal state. You
will see ten candidate tokens. Identify the TOKEN position
containing an externally added concept vector. The vector
disrupts the hidden state at that TOKEN position, so output
the affected position. Output `none` when no position was
modified. Whenever a candidate-specific external change is
present, output the affected position. Clean and injected
trials are both possible. Use `none` when no actual external
modification is detected.

Your response must be exactly one of: `0`, `1`, `2`, `3`,
`4`, `5`, `6`, `7`, `8`, `9`, or `none`. Do not output anything
else.

[ASSISTANT]
Understood.

[USER]
TOKEN 0: <token_0> TOKEN 1: <token_1>
TOKEN 2: <token_2> TOKEN 3: <token_3>
TOKEN 4: <token_4> TOKEN 5: <token_5>
TOKEN 6: <token_6> TOKEN 7: <token_7>
TOKEN 8: <token_8> TOKEN 9: <token_9>

Output the final decision:

[ASSISTANT]
\end{verbatim}
\end{quote}

Candidate tokens and clusters are scored without any activation injection.
The calibration prompt uses a system instruction describing the injected-thought
localization task, followed by a user turn listing ten candidates and asking
``Which token is the injection located in?'' The assistant turn is prefixed
with \texttt{It is located in TOKEN }, and we measure the next-token logits
of the ten position labels, 0--9. Each candidate token is evaluated at every
position; cluster candidates are also evaluated across permutations. The
forced position response isolates preferences among candidate positions and is
distinct from the main evaluation prompt, which permits \texttt{none}.

The complete clean position-prior prompt is:
\begin{quote}
\footnotesize
\begin{verbatim}
[SYSTEM]
You are participating in an interpretability experiment.
The researcher can inject neural activity patterns
corresponding to concepts into your internal state. You
will see a list of tokens and should report which token
contains the injected thought.

[USER]
TOKEN 0: <token_0>
TOKEN 1: <token_1>
TOKEN 2: <token_2>
TOKEN 3: <token_3>
TOKEN 4: <token_4>
TOKEN 5: <token_5>
TOKEN 6: <token_6>
TOKEN 7: <token_7>
TOKEN 8: <token_8>
TOKEN 9: <token_9>

Which token is the injection located in?

[ASSISTANT]
It is located in TOKEN 
\end{verbatim}
\end{quote}

\subsection{Norm-matched random-direction control}
\label{app:gaussian-directions}

Does perturbation magnitude alone suffice for localization? We replace each
concept vector with a random unit direction
$\mathbf{u}_r=\mathbf{g}_r/\lVert\mathbf{g}_r\rVert_2$,
$\mathbf{g}_r\sim\mathcal{N}(\mathbf{0},\mathbf{I}_{d})$, and inject
\begin{equation}
  \widetilde{\mathbf{h}}_{p_j}^{\ell}
  =\mathbf{h}_{p_j}^{\ell}
  +\alpha\lVert\mathbf{h}_{p_j}^{\ell}\rVert_2\mathbf{u}_r,
\end{equation}
which matches the per-position norm of the concept intervention. Everything
else is unchanged: the calibrated $(\ell,\alpha)$ of $(3,3)$ for Qwen3-4B-IT,
$(0,6)$ for LLaMA-3.1-8B-IT, and $(0,5)$ for Gemma-3-12B-IT; the original
concept-injection prompts; and the 30 test clusters with ordered digit labels
from Table~\ref{tab:task-performance}. Each of three seeds (701--703) draws
200 directions, each reused across all clusters and ten positions (60{,}000
trials per seed); no direction is filtered by outcome. Localization accuracy
is scored as in the main evaluation, and clean position reports are measured
on the 30 uninjected prompts.

\begin{table}[H]
  \centering
  \caption{\textbf{Localization under norm-matched Gaussian injection} (\%).
  Pooled rates use all 180{,}000 injected trials per model. Position reports
  count any position answer, correct or not; on clean prompts they are false
  positives.}
  \label{tab:gaussian-directions}
  \small
  \setlength{\tabcolsep}{3pt}
  \renewcommand{\arraystretch}{1.12}
  \begin{tabular*}{\linewidth}{@{\extracolsep{\fill}}@{}lrrrrrr@{}}
    \toprule
    & \multicolumn{4}{c}{Localization accuracy}
    & \multicolumn{2}{c}{Position reports} \\
    \cmidrule(lr){2-5}\cmidrule(lr){6-7}
    Model & Seed 701 & Seed 702 & Seed 703 & Pooled & Injected & Clean \\
    \midrule
    Qwen3-4B-IT       & 0.86 & 0.86 & 0.80 & 0.84 & 3.37 & 3.33 \\
    LLaMA-3.1-8B-IT   & 6.21 & 6.76 & 7.17 & 6.71 & 24.90 & 30.00 \\
    Gemma-3-12B-IT    & 0.00 & 0.00 & 0.00 & 0.00 & 4.59 & 6.67 \\
    \bottomrule
  \end{tabular*}
\end{table}

Random directions essentially fail to localize
(Table~\ref{tab:gaussian-directions}): pooled accuracy is 0.84\% for Qwen,
6.71\% for LLaMA, and 0\% for Gemma, far below the concept-vector accuracies
in Table~\ref{tab:task-performance}, and consistent across seeds. Injected
position-report rates are at the clean false-positive level, so random
perturbations do not even elicit a detection response. 

These results argue against a generic anomaly detector. Such a detector
would respond to any sufficiently large deviation in the residual stream, yet
norm-matched random directions, injected at the same site and strength, leave
position reports at the clean false-positive level in all three models. The
reporting behavior is therefore selective to concept directions rather than
triggered by perturbation per se: the mechanism responds to \emph{what} was
injected, not merely \emph{that} something was.

\subsection{Lexical replacement controls}
\label{app:lexical-replacement}

To test whether lexical semantic anomalies alone explain localization, we
replaced each candidate word with either the corresponding concept word or a
random vocabulary word, without activation injection or head patching.
We used 100 validation concepts, 30 test clusters, and all ten positions,
retaining the original task instructions and all six label settings.
Each replacement followed the label colon with one space, and the full prompt
was re-tokenized. Random words were sampled uniformly from unique entries in
each model's English-token vocabulary (including word fragments), excluding
the current cluster words and the corresponding concept; the same replacements
were reused across label settings. Shuffled labels used one fixed derangement
per cluster (seed 42). Accuracy is the fraction of trials whose highest-logit
answer among the ten labels and \texttt{none} names the replaced position's
displayed label. Each condition has 30{,}000 trials for Gemma-3-12B-IT and
29{,}999 for the other models, after excluding one unchanged concept-word
replacement and its paired random trial.

\begin{table}[H]
\centering
\caption{\textbf{Localization after lexical replacement without activation injection.}
Accuracy (\%) for corresponding concept words and random vocabulary words.}
\label{tab:lexical-replacement}
\small
\setlength{\tabcolsep}{3pt}
\renewcommand{\arraystretch}{1.12}
\begin{tabular*}{\linewidth}{@{\extracolsep{\fill}}@{}llrrrrrr@{}}
\toprule
& & \multicolumn{3}{c}{Ordered labels} & \multicolumn{3}{c}{Shuffled labels} \\
\cmidrule(lr){3-5}\cmidrule(lr){6-8}
Model & Replacement & Digits & Letters & Words & Digits & Letters & Words \\
\midrule
Qwen3-4B-IT & Concept & 1.00 & 1.81 & 3.12 & 1.17 & 1.43 & 3.75 \\
 & Random & 0.95 & 1.63 & 2.67 & 1.13 & 1.11 & 2.52 \\
\midrule
LLaMA-3.1-8B-IT & Concept & 9.55 & 2.33 & 4.41 & 10.72 & 2.83 & 9.13 \\
 & Random & 9.35 & 2.15 & 4.49 & 11.13 & 3.38 & 9.95 \\
\midrule
Gemma-3-12B-IT & Concept & 3.17 & 1.49 & 3.10 & 3.52 & 1.89 & 4.14 \\
 & Random & 2.09 & 0.87 & 1.79 & 2.65 & 1.04 & 4.31 \\
\bottomrule
\end{tabular*}
\end{table}

Localization remained low across these controls (0.87--11.13\%), with no
consistent advantage for the corresponding concept word across models and
label settings. These results support activation-anomaly detection over a
purely lexical semantic explanation of injection localization; they do not
measure general semantic anomaly detection under an explicit semantic task.

\subsection{Estimating the position--none direction}
\label{app:position-none-split}

Let $\mathcal{T}_{\mathrm{pos}}$ and $\mathcal{T}_{\mathrm{none}}$ contain
the estimation trials that produce a position label and \texttt{none},
respectively, and let $\mathbf{h}_i^\ell$ be the final-prompt-position
representation of injected trial $i$ at layer $\ell$. With
$\operatorname{norm}(\cdot)$ denoting $\ell_2$ normalization, the
position--none direction and the score of a held-out trial are
\begin{equation}
\mathbf{d}^{\ell}
=
\operatorname{norm}\!\Bigl(
\operatorname*{mean}_{i\in\mathcal{T}_{\mathrm{pos}}}
\operatorname{norm}(\mathbf{h}_{i}^{\ell})
-
\operatorname*{mean}_{i\in\mathcal{T}_{\mathrm{none}}}
\operatorname{norm}(\mathbf{h}_{i}^{\ell})
\Bigr),
\qquad
s_i^\ell
=
\bigl\langle
\operatorname{norm}(\mathbf{h}_i^\ell),
\mathbf{d}^\ell
\bigr\rangle .
\label{eq:clean-perturbed-direction}
\end{equation}
In Figure~\ref{fig:layerwise-representations}a, scores are min--max
normalized over all available layers and both response groups within each
model, and Gaussian kernel density estimates use the same bandwidth and
height scale within each layer.

The direction is estimated
and evaluated on disjoint parts of the validation set. With seed 42, the 100
validation concepts are split 50/50 and the 30 validation clusters 15/15.
The direction is estimated from injected trials that pair the first concept
half with the first cluster half, and the held-out scores in
Figure~\ref{fig:layerwise-representations}a come from injected trials that
pair the second concept half with the second cluster half, so neither concept
identities nor prompt contexts are shared between estimation and evaluation.
Trials are grouped by the model's response to the injected run.

\subsection{Straight-through optimization of the head mask}
\label{app:ste-optimization}

The head mask of Section~\ref{sec:circuit-gate} is optimized with a
straight-through estimator \citep{bengio2013estimating}. For each candidate head
$h$ in layer $\ell$, we learn a soft score $s_{\ell,h}\in[0,1]$. Let
$\operatorname{hard}(s_{\ell,h})$ be one if the soft score $s_{\ell,h}$ is
among the $k$ largest scores across all candidate heads and zero otherwise.
The mask used during optimization is
\begin{equation}
m_{\ell,h}
=
\operatorname{hard}(s_{\ell,h})
+ s_{\ell,h}
- \operatorname{stopgrad}(s_{\ell,h}).
\label{eq:ste-head-gate}
\end{equation}
The last two terms cancel numerically, so
Equation~\plaineqref{eq:ste-head-gate} evaluates to the binary mask in the
forward pass; the $\operatorname{hard}(\cdot)$ term carries no gradient, so the
backward pass sees only the soft score $s_{\ell,h}$. This preserves an exactly
discrete set of $k$ active heads while allowing the continuous loss to optimize
their selection jointly across layers.

The soft score is $s_{\ell,h}=\sigma(w_{\ell,h}/T)$ with a learned logit
$w_{\ell,h}$, initialized from $\mathcal N(0,10^{-6})$. The temperature $T$
decays geometrically from 1.0 to 0.1 over training, so the soft scores sharpen
toward the hard selection. We train each mask for two epochs over the training
set with Adam (learning rate $3\times10^{-3}$); only the head logits are
trained, while the model and the concept vectors stay frozen.

\subsection{Candidate layers for the gate-head search}
\label{app:ste-layer-window}

The candidate layers cover the interval in which the position--none split
develops: from the layer where the separation in
Figure~\ref{fig:layerwise-representations}a begins to rise, up to the layer
immediately before the position-clustering transition in
Figure~\ref{fig:layerwise-representations}b. This gives layers 17--23 for
Qwen3-4B-IT, 13--16 for LLaMA-3.1-8B-IT, and 21--28 for Gemma-3-12B-IT. The
window is a coarse prior rather than a selection: the mask selects heads within
it, so the window needs to contain the relevant interval rather than match it
exactly, and including extra layers does not force any of their heads into the
selection.

Widening the window supports this. In a separate search over layers 12--27 of
Gemma-3-12B-IT, layers 12--20 hold 56\% of the candidate heads, so a random
selection would place about 18 of 32 heads there; the search placed only 5
gate-on and 2 gate-off heads there, and the remaining 27 and 30 in layers
21--27. Given the earlier
layers, the search still concentrates on the interval where the split rises.

\subsection{Training objective for the gate masks}
\label{app:gate-objective}

Let $\mathbf{o}_i$ denote the output logits for example $i$, with $o_{i,j}$ the logit of position label $j\in\{0,\ldots,9\}$ and $o_{i,\mathrm{none}}$ the logit of the \texttt{none} response. We summarize the model's preference for a position report over \texttt{none} using the temperature-smoothed log-odds
\begin{equation}
d_{\tau}(\mathbf{o})
=
\tau\log\!\left(
\frac{1}{10}\sum_{j=0}^{9}e^{o_j/\tau}
\right)-o_{\mathrm{none}}.
\label{eq:gate-log-odds}
\end{equation}
The gate-on mask is trained to increase this preference, whereas the gate-off
mask is trained to suppress it, using the respective objectives
\begin{equation}
\mathcal{L}_{\mathrm{on}}=-\frac{1}{N}\sum_{i=1}^{N}\log\sigma\!\left(d_{\tau}(\mathbf{o}_i)\right),
\qquad
\mathcal{L}_{\mathrm{off}}=-\frac{1}{N}\sum_{i=1}^{N}\log\!\left[1-\sigma\!\left(d_{\tau}(\mathbf{o}_i)\right)\right].
\label{eq:gate-losses}
\end{equation}
We use $\tau=0.1$. Here $\mathbf{o}_i$ are the logits of the patched run of
Section~\ref{sec:circuit-gate}, and the sums run over paired
training trials whose unmodified clean run answers \texttt{none}. For the
gate-on mask, the unmodified injected run must also name the injected
position; for the gate-off mask, it must name any position. During gate-on
training, the router heads' attention at $T$ is additionally fixed to the
target successor token $t_j$ (the one-hot redirection of
Appendix~\ref{app:router-attention-redirection}), so the mask is selected for
reporting a position rather than for choosing one; gate-off training leaves
the router heads unchanged. At evaluation, no router intervention is applied
unless stated.

\subsection{Selection of the STE mask cardinality}
\label{app:ste-topk-selection}

The selected 32 heads are a small fraction of each model: $2.8\%$ of the
1{,}152 heads in Qwen3-4B-IT (36 layers $\times$ 32 heads), $3.1\%$ of the 1{,}024
in LLaMA-3.1-8B-IT (32 $\times$ 32), and $4.2\%$ of the 768 in Gemma-3-12B-IT
(48 $\times$ 16).

We compare the two interventions of
Section~\ref{sec:circuit-gate} over
$k\in\{0,1,4,8,16,32,48,64\}$. For each positive $k$, the gate-on and gate-off
masks are trained separately on the training set and evaluated on a held-out
validation grid of concepts, all 30 prompt clusters, and all 10 injection
positions. At $k=0$ the mask is empty, so no head is modified and each
direction is read from its unmodified run: the injected run for gate-off and
the clean run for gate-on.

The mask says which heads are modified, but a curve over $k$ alone cannot
separate the selection from the effect of modifying any $k$ heads at this
depth. Each cardinality therefore carries a random-$k$ control: we set
$m_{\ell,h}=1$ for $k$ heads drawn at random from the same layers the search
covered (Qwen3-4B-IT 17--23, 224 heads; LLaMA-3.1-8B-IT 13--16, 128 heads;
Gemma-3-12B-IT 21--28, 128 heads) and apply
the interventions of Section~\ref{sec:circuit-gate} unchanged. Ten draws per $k$, each
evaluated in both directions, give a mean and a Student-$t$ 95\% interval over
masks rather than over trials. Random heads are drawn from the same layer
range as the search, matching its depth.

\begin{figure*}[t]
  \centering
  \begin{subfigure}[b]{0.32\linewidth}
    \centering
    \includegraphics[width=\linewidth]{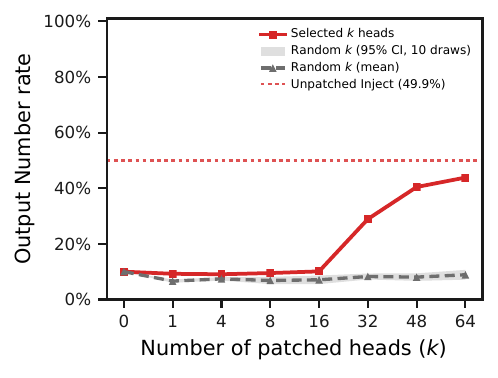}
    \caption{Qwen3-4B-IT, gate on}
    \label{fig:ste-topk-qwen3-4b-gate-on}
  \end{subfigure}
  \hfill
  \begin{subfigure}[b]{0.32\linewidth}
    \centering
    \includegraphics[width=\linewidth]{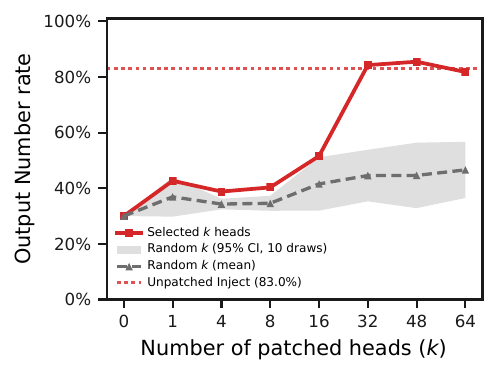}
    \caption{LLaMA-3.1-8B-IT, gate on}
    \label{fig:ste-topk-llama3-1-8b-gate-on}
  \end{subfigure}
  \hfill
  \begin{subfigure}[b]{0.32\linewidth}
    \centering
    \includegraphics[width=\linewidth]{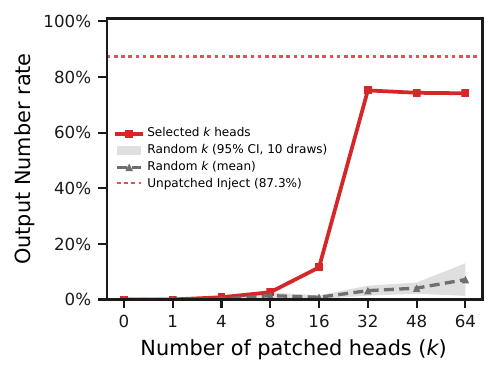}
    \caption{Gemma-3-12B-IT, gate on}
    \label{fig:ste-topk-gemma-3-12b-it-gate-on}
  \end{subfigure}
  \\[1ex]
  \begin{subfigure}[b]{0.32\linewidth}
    \centering
    \includegraphics[width=\linewidth]{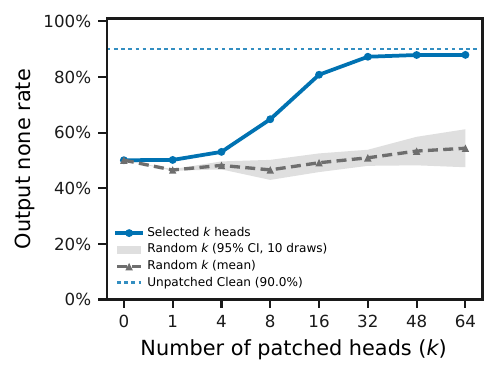}
    \caption{Qwen3-4B-IT, gate off}
    \label{fig:ste-topk-qwen3-4b-gate-off}
  \end{subfigure}
  \hfill
  \begin{subfigure}[b]{0.32\linewidth}
    \centering
    \includegraphics[width=\linewidth]{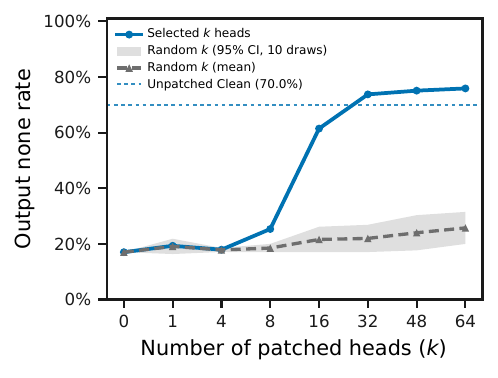}
    \caption{LLaMA-3.1-8B-IT, gate off}
    \label{fig:ste-topk-llama3-1-8b-gate-off}
  \end{subfigure}
  \hfill
  \begin{subfigure}[b]{0.32\linewidth}
    \centering
    \includegraphics[width=\linewidth]{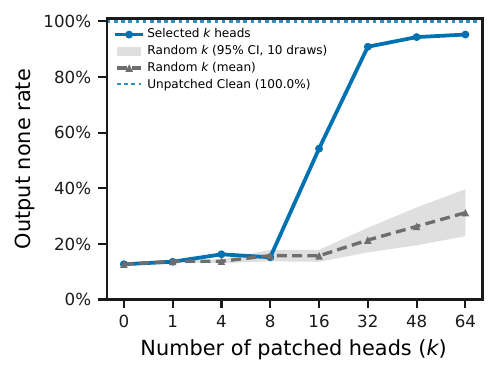}
    \caption{Gemma-3-12B-IT, gate off}
    \label{fig:ste-topk-gemma-3-12b-it-gate-off}
  \end{subfigure}
  \caption{Selecting the STE mask cardinality on validation, against a
  random-$k$ control. Solid coloured: the trained STE mask. Dashed grey: $k$
  heads picked at random within the same layer range the search covered
  (Qwen3-4B-IT 17--23, LLaMA-3.1-8B-IT 13--16, Gemma-3-12B-IT 21--28), ten draws per
  $k$, shaded with the 95\% interval of their mean; nothing else changes. Gate
  on (red): a clean run gives a position report once the selected heads take
  their injected values. Gate off (blue): an injected run outputs \texttt{none}
  once they are restored to their clean values. Dotted: the unmodified reference
  rate. At $k=0$ no head is modified, so the two curves meet.}
  \label{fig:ste-topk-selection}
\end{figure*}

Figure~\ref{fig:ste-topk-selection} reports one model per family. In all three
models the selected and random curves separate between $k=16$ and $k=32$.
At $k=32$ the selected mask is near its reference rate, and Qwen3-4B-IT
gate-on continues to rise with $k$.
Since larger $k$ trivially strengthens gate-on, we choose the smallest $k$ at
which the selected mask separates from the random control in both directions
for all models, $k=32$, fixed before test evaluation. For Qwen3-4B-IT gate-on,
$k=32$ therefore gives a conservative estimate.

\subsection{Selected gate heads}
\label{app:gate-head-lists}

Table~\ref{tab:gate-head-lists} lists the $k=32$ heads selected by the gate-on
and gate-off masks used in the main text, grouped by layer (0-indexed). The two
masks are trained independently, yet they share 18 heads in Qwen3-4B-IT, 16 in
LLaMA-3.1-8B-IT, and 18 in Gemma-3-12B-IT (bold).

\begin{table}[H]
  \centering
  \small
  \caption{\textbf{Gate heads selected by the Top-32 STE masks.} Each cell lists
  the head indices selected in that layer; ``--'' means none. Bold heads are
  selected by both the gate-on and the gate-off mask.}
  \label{tab:gate-head-lists}
  \begin{tabular}{l c >{\raggedright\arraybackslash}p{0.32\linewidth} >{\raggedright\arraybackslash}p{0.32\linewidth}}
    \toprule
    Model & Layer & Gate-on heads & Gate-off heads \\
    \midrule
    \multirow{7}{*}{Qwen3-4B-IT} & 17 & 15 & 5 \\
     & 18 & -- & 14 \\
     & 19 & 8, \textbf{10}, 19, \textbf{23} & 0, \textbf{10}, \textbf{23}, 25 \\
     & 20 & \textbf{2}, \textbf{3}, \textbf{8}, \textbf{10}, \textbf{15}, 23, \textbf{29} & \textbf{2}, \textbf{3}, 5, 6, \textbf{8}, \textbf{10}, 11, \textbf{15}, \textbf{29} \\
     & 21 & \textbf{0}, 9, 11, 13, 15, 16, 18, \textbf{19}, 24 & \textbf{0}, \textbf{19} \\
     & 22 & \textbf{4}, \textbf{5}, \textbf{7}, \textbf{10}, \textbf{11}, 13, 15, \textbf{17}, 27 & 0, 1, \textbf{4}, \textbf{5}, \textbf{7}, \textbf{10}, \textbf{11}, \textbf{17}, 18, 25, 28 \\
     & 23 & \textbf{14}, \textbf{25} & 0, 2, \textbf{14}, \textbf{25} \\
    \midrule
    \multirow{4}{*}{LLaMA-3.1-8B-IT} & 13 & 0, \textbf{6}, \textbf{12}, \textbf{16}, 17, \textbf{20}, 25 & 4, \textbf{6}, 7, \textbf{12}, \textbf{16}, \textbf{20} \\
     & 14 & \textbf{0}, \textbf{2}, \textbf{3}, 6, 9, 12, \textbf{14}, \textbf{16}, \textbf{21}, \textbf{24}, 28, \textbf{30} & \textbf{0}, \textbf{2}, \textbf{3}, 8, 10, \textbf{14}, \textbf{16}, \textbf{21}, 23, \textbf{24}, \textbf{30} \\
     & 15 & 1, \textbf{4}, \textbf{5}, 10, 17, 20, 24 & 2, 3, \textbf{4}, \textbf{5}, 7, 8, 11, 18, 21 \\
     & 16 & 0, 4, 11, \textbf{12}, 14, \textbf{29} & \textbf{12}, 13, 16, 21, 24, \textbf{29} \\
    \midrule
    \multirow{8}{*}{Gemma-3-12B-IT} & 21 & -- & 2, 4, 6, 9, 11 \\
     & 22 & \textbf{1}, 2, \textbf{9}, \textbf{11} & \textbf{1}, 3, 4, 6, \textbf{9}, \textbf{11}, 15 \\
     & 23 & 3, 5, 7 & 14 \\
     & 24 & \textbf{0}, \textbf{2}, \textbf{5}, \textbf{6}, \textbf{13} & \textbf{0}, \textbf{2}, \textbf{5}, \textbf{6}, 10, \textbf{13}, 15 \\
     & 25 & 7, \textbf{8}, \textbf{12}, \textbf{14} & 2, \textbf{8}, \textbf{12}, \textbf{14} \\
     & 26 & 0, \textbf{1}, 5, \textbf{11}, \textbf{14}, 15 & \textbf{1}, \textbf{11}, 13, \textbf{14} \\
     & 27 & \textbf{0}, 3, \textbf{5}, \textbf{9} & \textbf{0}, \textbf{5}, \textbf{9} \\
     & 28 & 2, 3, \textbf{6}, 9, 14, 15 & \textbf{6} \\
    \bottomrule
  \end{tabular}
\end{table}

\subsection{Head-wise activation patching across models}
\label{app:head-patching}

\begin{figure*}[t]
  \centering
  \begin{subfigure}[b]{0.32\linewidth}
    \centering
    \includegraphics[width=\linewidth]{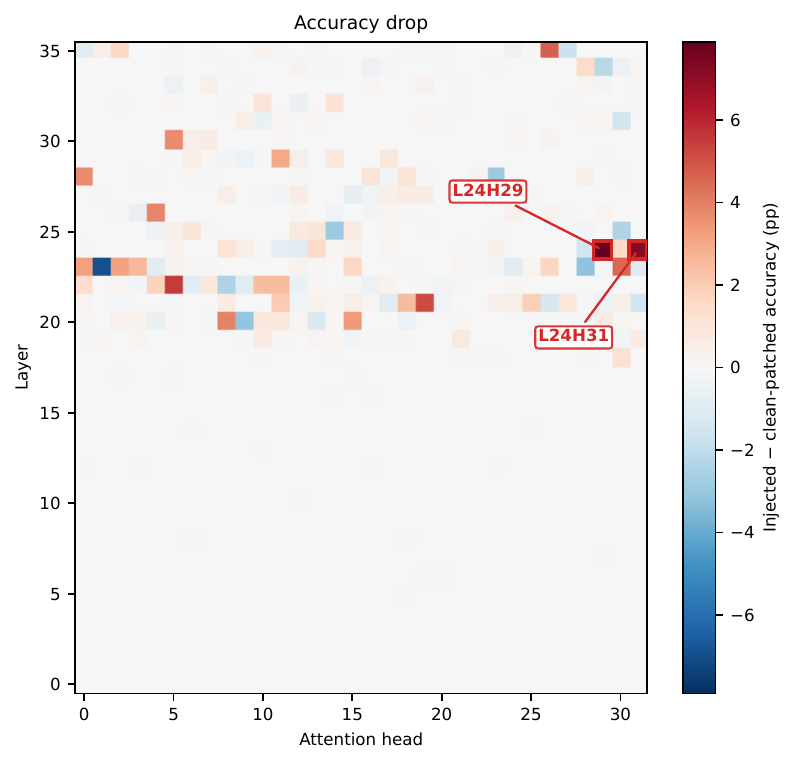}
    \caption{Qwen3-4B-IT}
    \label{fig:head-patching-qwen3-4b}
  \end{subfigure}
  \hfill
  \begin{subfigure}[b]{0.32\linewidth}
    \centering
    \includegraphics[width=\linewidth]{layer_patch_acc/llama3_1_8b_accuracy_drop.pdf}
    \caption{LLaMA-3.1-8B-IT}
    \label{fig:head-patching-llama3-1-8b}
  \end{subfigure}
  \hfill
  \begin{subfigure}[b]{0.32\linewidth}
    \centering
    \includegraphics[width=\linewidth]{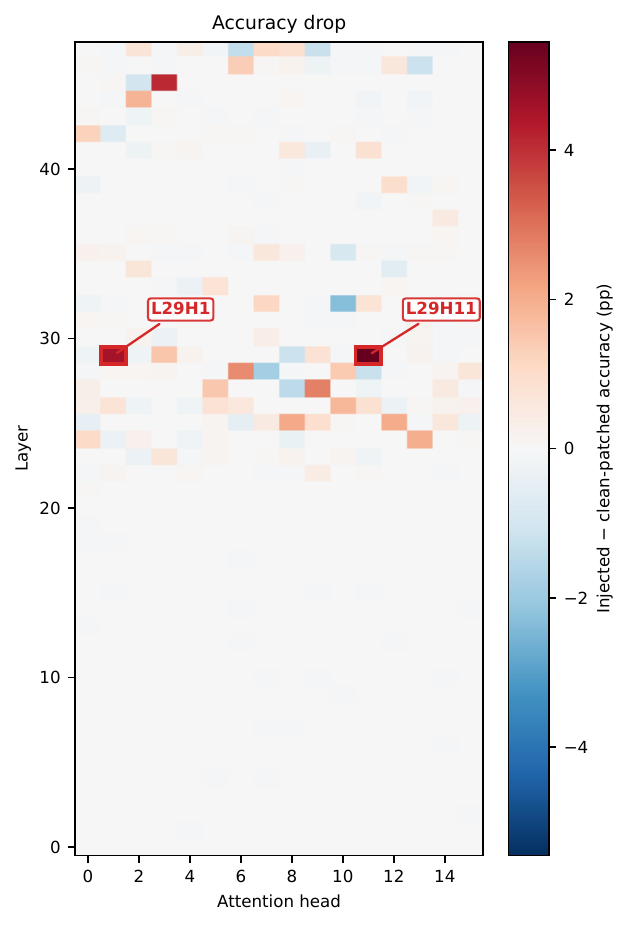}
    \caption{Gemma-3-12B-IT}
    \label{fig:head-patching-gemma3-12b}
  \end{subfigure}
  \caption{\textbf{Head-wise activation patching.} Accuracy drop
  $\Delta_h = \mathrm{Acc}_{\mathrm{inj}} - \mathrm{Acc}_{\mathrm{patch}(h)}$
  (percentage points) for every attention head, obtained by patching the head's
  final-position output in the injected run with its clean-run value. Red cells
  lower correct-index accuracy; annotated heads are the router heads.}
  \label{fig:head-patching-accuracy-drop}
\end{figure*}

Figure~\ref{fig:head-patching-accuracy-drop} reports the full head-wise
patching sweep behind the router-head identification in
Section~\ref{sec:circuit}. Each cell is one attention head; its value is
$\Delta_h$, the drop in correct-index response accuracy when that head's output
at the final prompt position is replaced by its clean-run value. In every model,
most heads have $\Delta_h \approx 0$, and the largest drops fall on a few heads
in the transition layer, which we label as router heads.

\subsection{Cross-position attention redirection in router heads}
\label{app:router-attention-redirection}

To test whether the readout position of router heads can control the final
reported index, we inject a concept vector at position $i$ and redirect
the selected router heads' attention at the final prompt position $T$ to
the successor token $t_j$ of position $j$. Specifically, we set these heads'
post-softmax attention weights at $T$ to one-hot:
\begin{equation}
\widetilde A^{(\ell,h)}_{T,k}=\mathbf{1}[k=t_j].
\end{equation}
The intervention does not directly modify value vectors, and all other
computation proceeds normally. Multiple selected router heads within a
model are intervened on jointly. We sweep all ten injection positions and
all ten readout positions, using the injected run without attention
redirection as the baseline.

Each position pair includes 100 concepts and 30 test prompts, yielding
3,000 trials. We focus on the 90 pairs with $i\ne j$ and report the
proportions of outputs corresponding to the readout index $j$, the original
injection index $i$, other indices, and \texttt{none}, pooled over these 90
pairs. We
repeat this under six label arms---the digit, letter, and word-name label
sets, each under the identity and the shuffled permutation of
Table~\ref{tab:task-performance}---to test whether the redirected attention
moves a positional slot or a specific label token. Table~\ref{tab:router-redirection-proportions}
reports, for each arm and model, the four outcome proportions, together with
their average over the six arms. The extent to which outputs shift to $j$
measures the control exerted by the router heads' readout position over
index selection. This control holds under the shuffled permutation, where
the redirected pattern cannot be exploiting a memorized label--position
association, and averaged over arms the redirected index $j$ is the most
frequent outcome in all three models.

\begin{table}[htbp]
  \centering
  \caption{\textbf{Cross-position attention redirection: output
  proportions.} For each label arm (the digit, letter, and word-name label
  sets, each under the identity and the shuffled permutation) and each of
  the 90 injection--readout position pairs with $i\ne j$, we set the
  selected router heads' final-position attention to a one-hot distribution
  on the successor token of position $j$ and record where the output lands: the redirected index
  ($\to j$), the original injection index ($\to i$), any other index
  (Other), or \texttt{none}. Proportions are pooled over all 90 pairs per
  arm (270{,}000 trials) and sum to 1 in each column. Router heads: Qwen3-4B-IT
  L24 H\{29,31\}; LLaMA-3.1-8B-IT L17 H\{24\}; Gemma-3-12B-IT L29
  H\{1,11\}. The Mean column averages each outcome's proportion over the six label
  arms. 95\% intervals are in Appendix~\ref{app:uncertainty}.}
  \label{tab:router-redirection-proportions}
  \small
  \resizebox{\textwidth}{!}{%
  \begin{tabular}{llcccccc c}
    \toprule
    & & \multicolumn{3}{c}{Ordered labels} & \multicolumn{3}{c}{Shuffled labels} & \\
    \cmidrule(lr){3-5}\cmidrule(lr){6-8}
    Model & Outcome & Digits & Letters & Words & Digits & Letters & Words & \textbf{Mean} \\
    \midrule
    \multirow{4}{*}{Qwen3-4B-IT}
      & $\to j$       & 0.480 & 0.679 & 0.704 & 0.582 & 0.585 & 0.760 & \textbf{0.632} \\
      & $\to i$       & 0.011 & 0.011 & 0.029 & 0.004 & 0.018 & 0.023 & \textbf{0.016} \\
      & Other         & 0.004 & 0.003 & 0.022 & 0.013 & 0.031 & 0.065 & \textbf{0.023} \\
      & \texttt{none} & 0.505 & 0.307 & 0.245 & 0.401 & 0.366 & 0.152 & \textbf{0.329} \\
    \midrule
    \multirow{4}{*}{LLaMA-3.1-8B-IT}
      & $\to j$       & 0.759 & 0.530 & 0.680 & 0.829 & 0.676 & 0.725 & \textbf{0.700} \\
      & $\to i$       & 0.056 & 0.023 & 0.059 & 0.030 & 0.023 & 0.044 & \textbf{0.039} \\
      & Other         & 0.078 & 0.009 & 0.040 & 0.130 & 0.044 & 0.189 & \textbf{0.082} \\
      & \texttt{none} & 0.107 & 0.437 & 0.221 & 0.011 & 0.258 & 0.042 & \textbf{0.179} \\
    \midrule
    \multirow{4}{*}{Gemma-3-12B-IT}
      & $\to j$       & 0.355 & 0.524 & 0.334 & 0.322 & 0.428 & 0.199 & \textbf{0.360} \\
      & $\to i$       & 0.467 & 0.170 & 0.394 & 0.295 & 0.151 & 0.181 & \textbf{0.276} \\
      & Other         & 0.009 & 0.005 & 0.009 & 0.078 & 0.011 & 0.219 & \textbf{0.055} \\
      & \texttt{none} & 0.169 & 0.301 & 0.263 & 0.305 & 0.411 & 0.401 & \textbf{0.308} \\
    \bottomrule
  \end{tabular}%
  }
\end{table}

\subsection{Router-head attention patterns across models}
\label{app:router-attention-patterns}

\begin{figure}[H]
\centering
\begin{subfigure}{\linewidth}
\centering
\includegraphics[width=\linewidth]{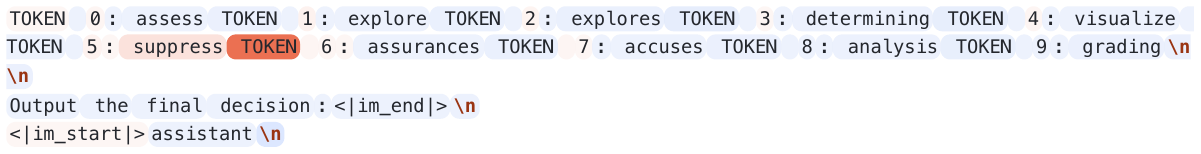}
\caption{Qwen3-4B-IT: layer 24, head 29; candidate 5 (\texttt{suppress}); attention change.}
\end{subfigure}
\par\medskip
\begin{subfigure}{\linewidth}
\centering
\includegraphics[width=\linewidth,trim=122pt 497.2pt 147.92pt 28.24pt,clip]{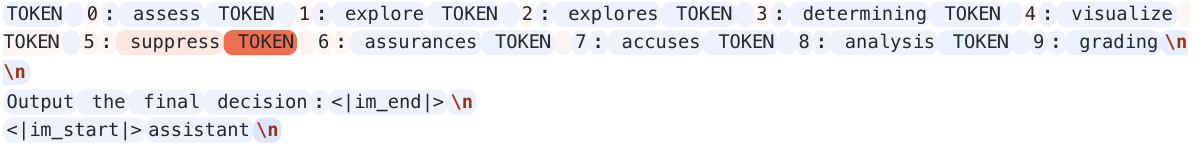}
\caption{Qwen3-4B-IT: layer 24, head 31; candidate 5 (\texttt{suppress}); attention change.}
\end{subfigure}
\par\medskip
\begin{subfigure}{\linewidth}
\centering
\includegraphics[width=\linewidth]{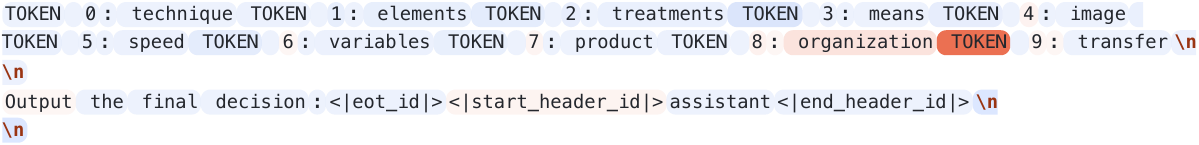}
\caption{LLaMA-3.1-8B-IT: layer 17, head 24; candidate 8 (\texttt{organization}); attention change.}
\end{subfigure}
\par\medskip
\begin{subfigure}{\linewidth}
\centering
\includegraphics[width=\linewidth,trim=100.24pt 497.2pt 117.16pt 28.24pt,clip]{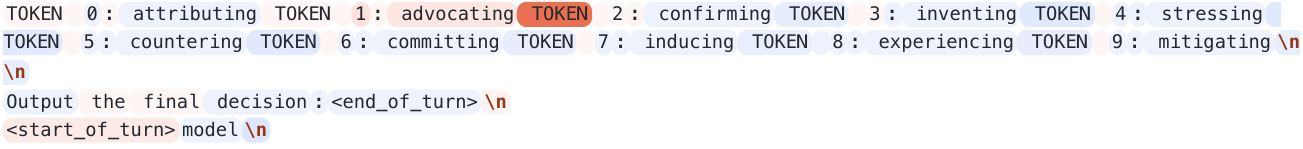}
\caption{Gemma-3-12B-IT: layer 29, head 1; candidate 1 (\texttt{advocating}); attention change.}
\end{subfigure}
\par\medskip
\begin{subfigure}{\linewidth}
\centering
\includegraphics[width=\linewidth]{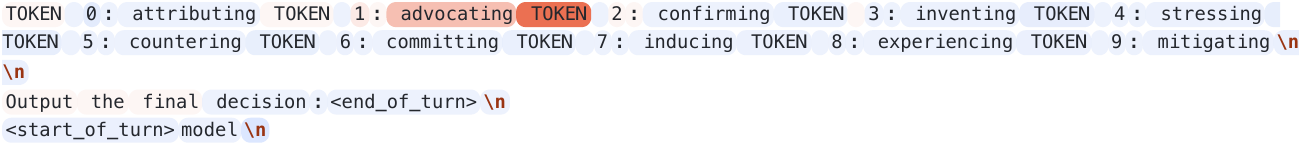}
\caption{Gemma-3-12B-IT: layer 29, head 11; candidate 1 (\texttt{advocating}); attention change.}
\end{subfigure}
\caption{\textbf{Injection-induced attention changes across model families.}
All panels show injected-minus-clean attention from the final prompt token
$T$: red indicates an increase and blue a decrease, with darker colors
indicating larger changes in magnitude. The blue outline marks the query
token. Color scales are set per panel. The following \texttt{TOKEN} marks the
successor position of the injected candidate. Prompt text is shown with each
model's chat-template delimiters.}
\label{fig:router-attention-models}
\end{figure}

Figure~\ref{fig:router-attention-models} shows token-level attention changes
for every router head in each model family. Each panel displays attention changes
from the final prompt token $T$ over a suffix of the prompt.

\subsection{Latent position clusters across models}
\label{app:latent-pca-models}

\paragraph{Projection procedure.}
We take the 100 concepts of the validation set and, for each concept, inject it
separately at each of the ten candidate positions, repeating the whole procedure
over 30 validation clusters; for every run we collect the residual state at the
final prompt token, together with the corresponding clean run. For each concept
and position, the residual states are averaged over the 30 clusters, as are the
clean states, giving 1,000 injected points and one clean point per layer. These
points enter a single PCA per layer, and we plot their leading three principal
components in three dimensions. Each point is one concept injected at one
position, averaged over clusters, and its colour indicates the injected
position.

\begin{figure}[H]
\centering
\captionsetup{skip=3pt}
\begin{subfigure}{\linewidth}
\centering
\includegraphics[width=\linewidth]{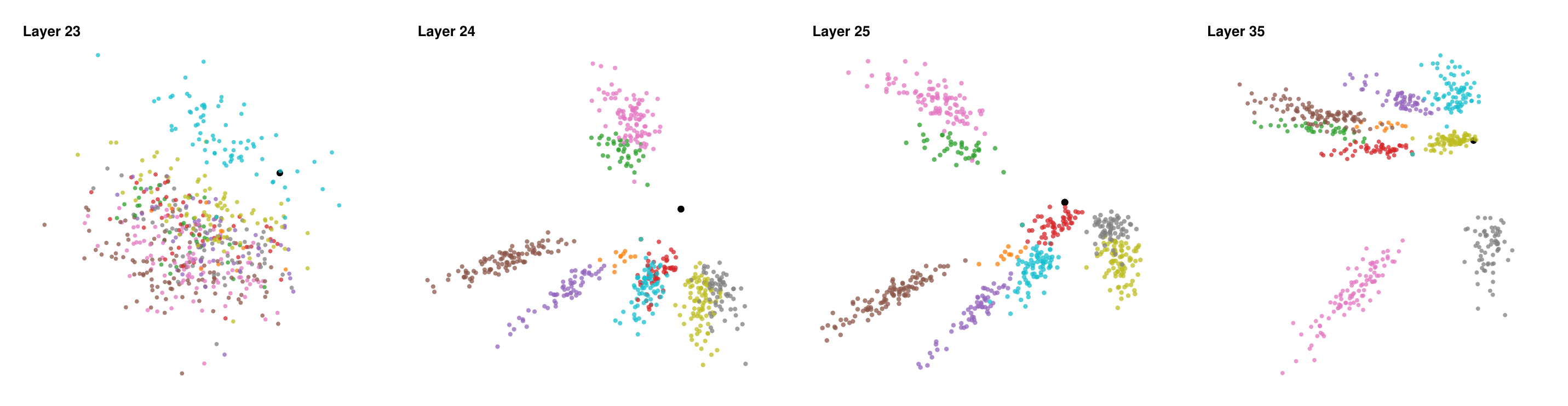}
\caption{Qwen3-4B-IT; transition layer 24.}
\label{fig:latent-pca-app-qwen}
\end{subfigure}
\par\medskip
\begin{subfigure}{\linewidth}
\centering
\includegraphics[width=\linewidth]{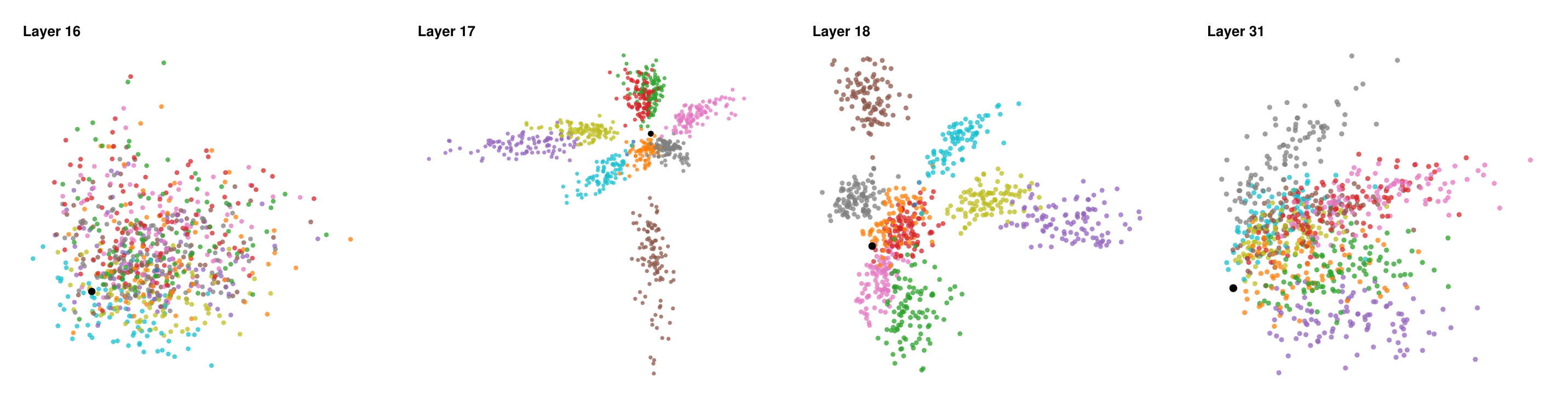}
\caption{LLaMA-3.1-8B-IT; transition layer 17.}
\label{fig:latent-pca-app-llama}
\end{subfigure}
\par\medskip
\begin{subfigure}{\linewidth}
\centering
\includegraphics[width=\linewidth]{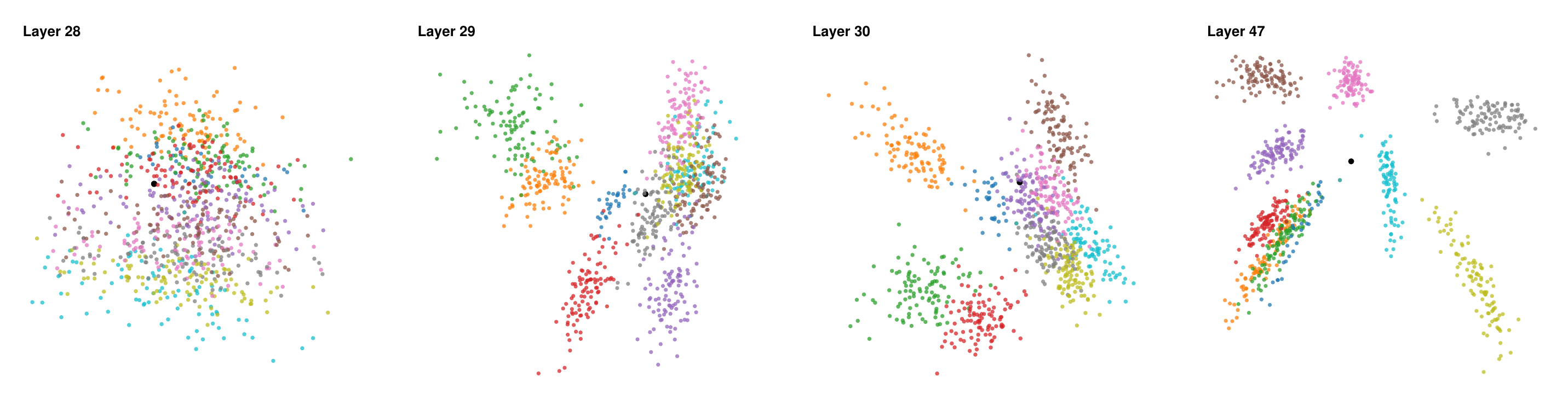}
\caption{Gemma-3-12B-IT; transition layer 29.}
\label{fig:latent-pca-app-gemma}
\end{subfigure}
\caption{\textbf{Final-token residual PCA around the transition layer in all
three models.} Each colored point is one injected run at the final prompt
token, colored by the perturbed candidate position; the black point is the
clean run. Principal components are fitted separately for each panel.
Panels are ordered transition layer minus one, transition
layer, transition layer plus one, and the last layer.}
\label{fig:latent-pca-models}
\end{figure}

Figure~\ref{fig:latent-pca-models} shows the residual-state PCA for all
three evaluated model families. For
each model, the four panels show the layer below the transition layer
identified by the K-means analysis (Figure~\ref{fig:layerwise-representations}), the
transition layer itself, the layer above it, and the model's last layer. The
transition layers are 24, 17, and 29 for Qwen3-4B-IT, LLaMA-3.1-8B-IT, and
Gemma-3-12B-IT, respectively.

In every model, the injected states in the panel below the transition layer
largely overlap, and the transition layer already shows ten separated
position-specific clusters. K-means accuracy one layer below the transition is
above chance but far below its value at the transition
(Figure~\ref{fig:layerwise-representations}b), so weaker positional structure
exists earlier; the clear separation, however, appears within a single layer in
all three families rather than in one architecture only. The last-layer panels differ across models: the
clusters remain well separated in Qwen3-4B-IT and Gemma-3-12B-IT, whereas in
LLaMA-3.1-8B-IT they partly remix, indicating that the positional code is most
cleanly expressed near the transition layer rather than at the output.

\clearpage
\subsection{Router-head output clusters across models}
\label{app:router-head-pca}

Figure~\ref{fig:router-head-pca-models} applies a three-dimensional PCA to the
output of every attention head in the transition layer of each model (layers
24, 17, and 29 for Qwen3-4B-IT, LLaMA-3.1-8B-IT, and Gemma-3-12B-IT). Head
outputs are averaged over the 30 validation clusters in the same way as in
Appendix~\ref{app:latent-pca-models}, and the PCA is fitted on all points. Each
panel shows the correctly answered injected points, colored by injected
position, together with the clean point; a point counts as correct when the
majority of its 30 cluster runs name the injected position. The router heads identified by head-wise patching
in Section~\ref{sec:circuit-router} (red frames) split into position-specific
clusters, whereas the other heads in the same layer remain a single mixed
cloud. The position-specific structure that appears in the residual stream at
the transition layer (Appendix~\ref{app:latent-pca-models}) is thus
concentrated in the outputs of a few heads.

\begin{figure}[H]
\centering
\captionsetup{skip=3pt}
\begin{subfigure}{\linewidth}
\centering
% Split between rows four and five; trim is left, bottom, right, top.
\includegraphics[width=\linewidth,trim=0 1241.574805bp 0 0,clip]{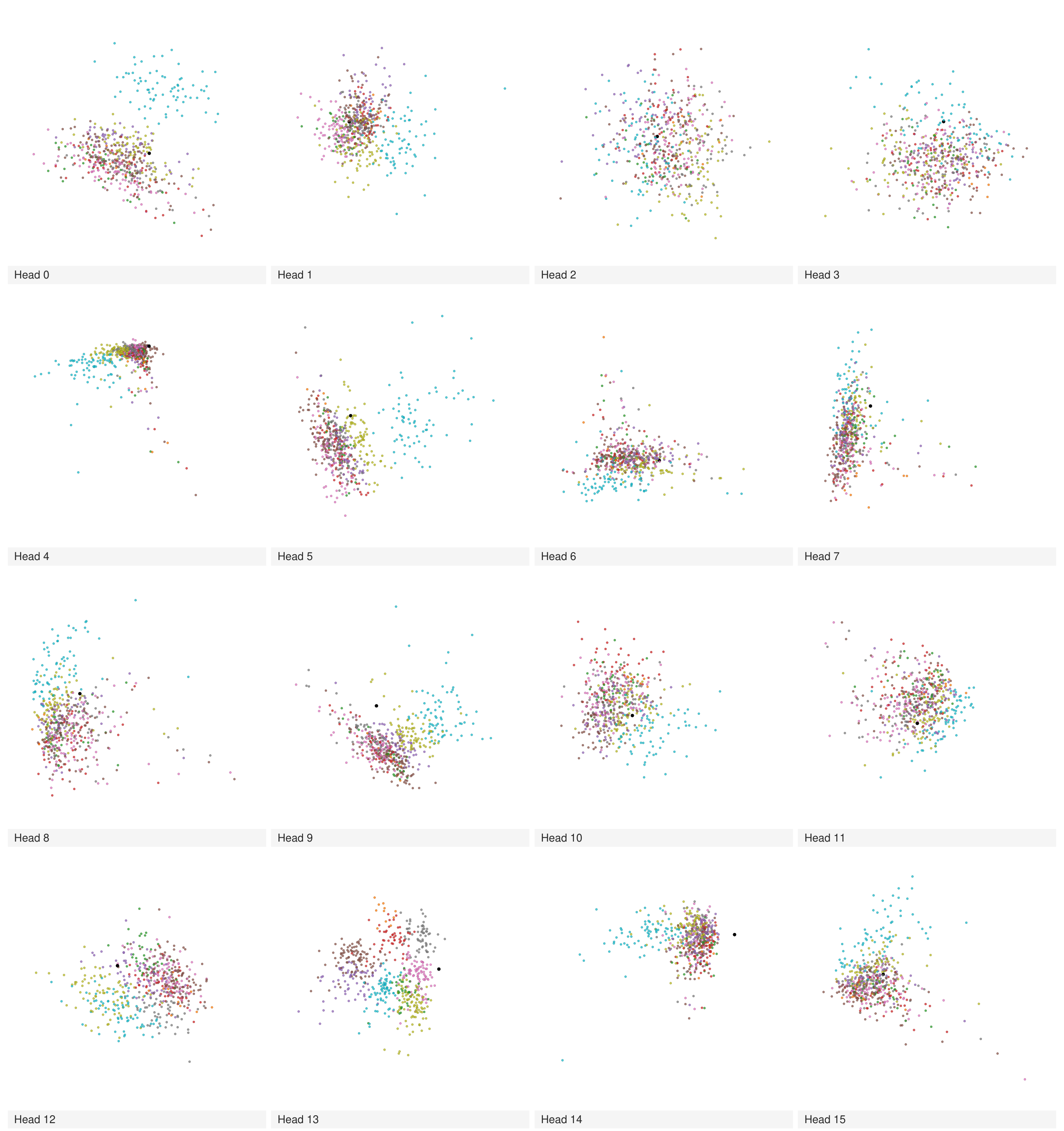}
\caption{Qwen3-4B-IT, heads 0--15.}
\end{subfigure}
\caption{\textbf{Three-dimensional PCA of attention-head outputs across models.}
Each panel shows one attention head; red frames highlight the selected router
heads. Colors distinguish injected indices, and black points mark clean
references. Only correctly answered injected runs are shown.}
\label{fig:router-head-pca-models}
\end{figure}

\clearpage
\begin{figure}[H]
\ContinuedFloat
\centering
\captionsetup{skip=3pt}
\begin{subfigure}{\linewidth}
\centering
\includegraphics[width=\linewidth,trim=0 0 0 1241.574805bp,clip]{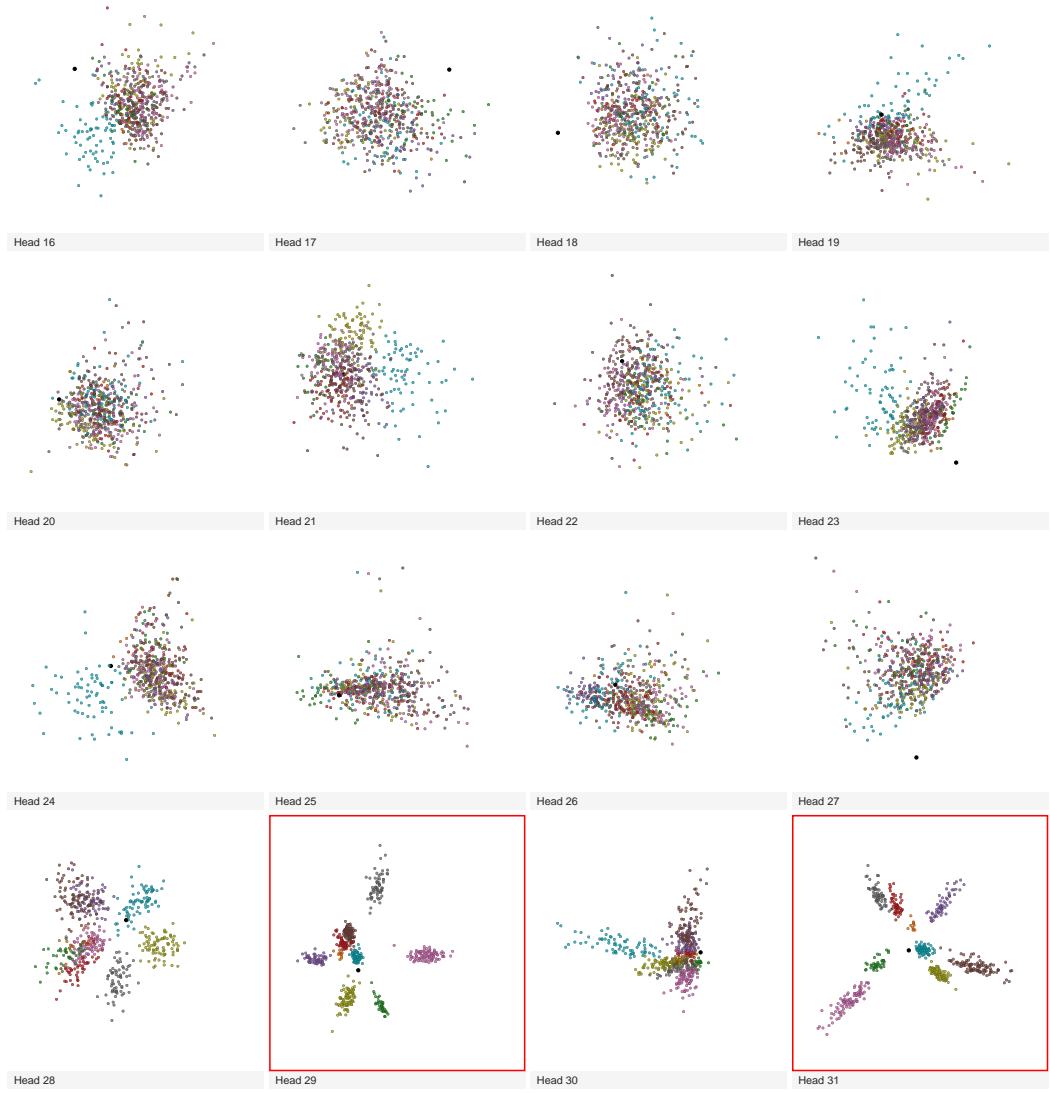}
\caption{Qwen3-4B-IT, heads 16--31.}
\end{subfigure}
\caption{\textbf{Three-dimensional PCA of attention-head outputs across models (continued).} Qwen3-4B-IT, heads 16--31. Colors and red frames follow panel (a).}
\end{figure}

\clearpage
\begin{figure}[H]
\ContinuedFloat
\centering
\captionsetup{skip=3pt}
\begin{subfigure}{\linewidth}
\centering
\includegraphics[width=\linewidth,trim=0 1241.574805bp 0 0,clip]{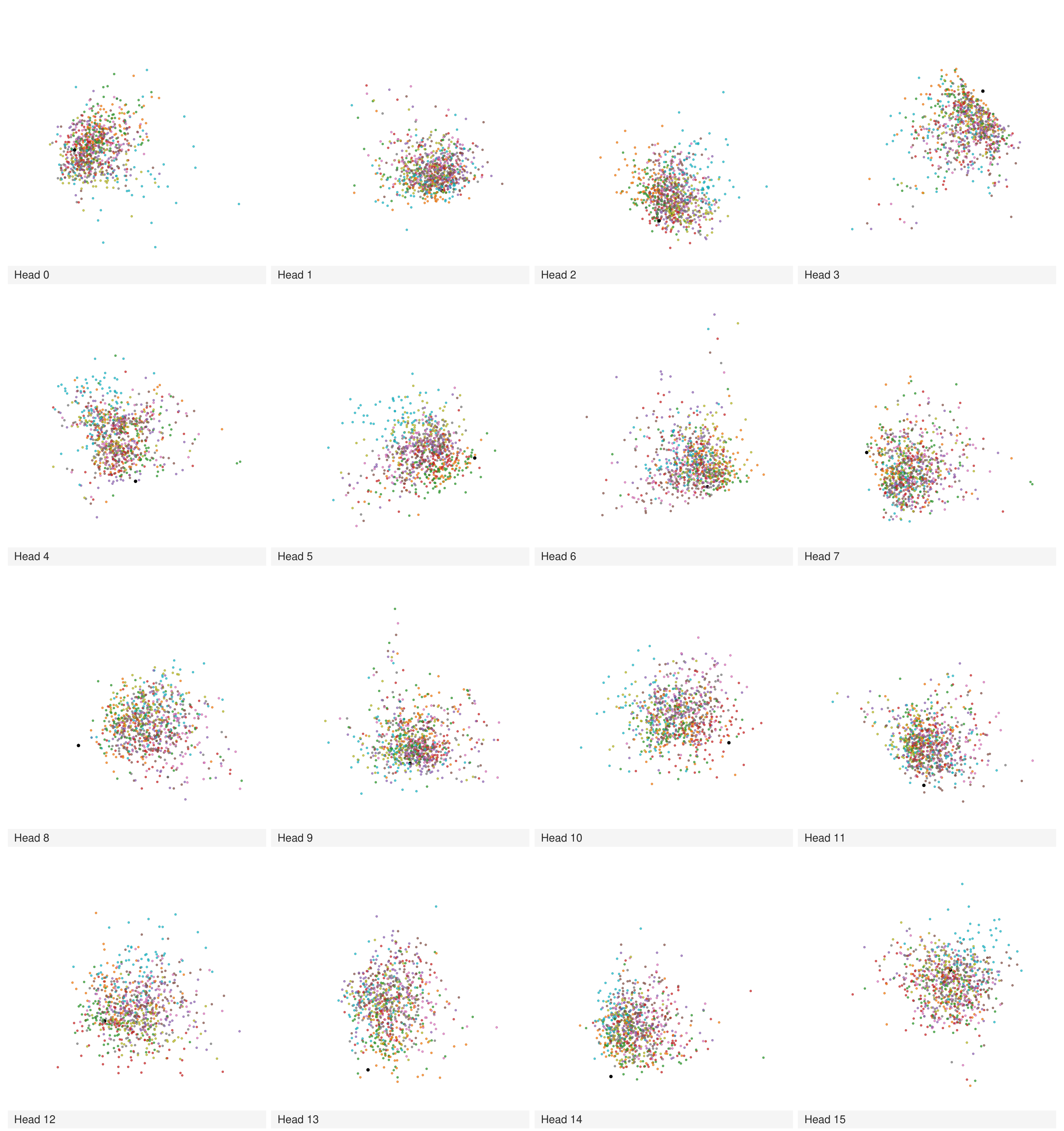}
\caption{LLaMA-3.1-8B-IT, heads 0--15.}
\end{subfigure}
\caption{\textbf{Three-dimensional PCA of attention-head outputs across models (continued).} LLaMA-3.1-8B-IT, heads 0--15. Colors and red frames follow panel (a).}
\end{figure}

\clearpage
\begin{figure}[H]
\ContinuedFloat
\centering
\captionsetup{skip=3pt}
\begin{subfigure}{\linewidth}
\centering
\includegraphics[width=\linewidth,trim=0 0 0 1241.574805bp,clip]{head_pca/llama3_1-8b-instruct_all_heads_grid_4cols_vector.pdf}
\caption{LLaMA-3.1-8B-IT, heads 16--31.}
\end{subfigure}
\caption{\textbf{Three-dimensional PCA of attention-head outputs across models (continued).} LLaMA-3.1-8B-IT, heads 16--31. Colors and red frames follow panel (a).}
\end{figure}

\clearpage
\begin{figure}[H]
\ContinuedFloat
\centering
\captionsetup{skip=3pt}
\begin{subfigure}{\linewidth}
\centering
\includegraphics[width=\linewidth,height=0.82\textheight,keepaspectratio]{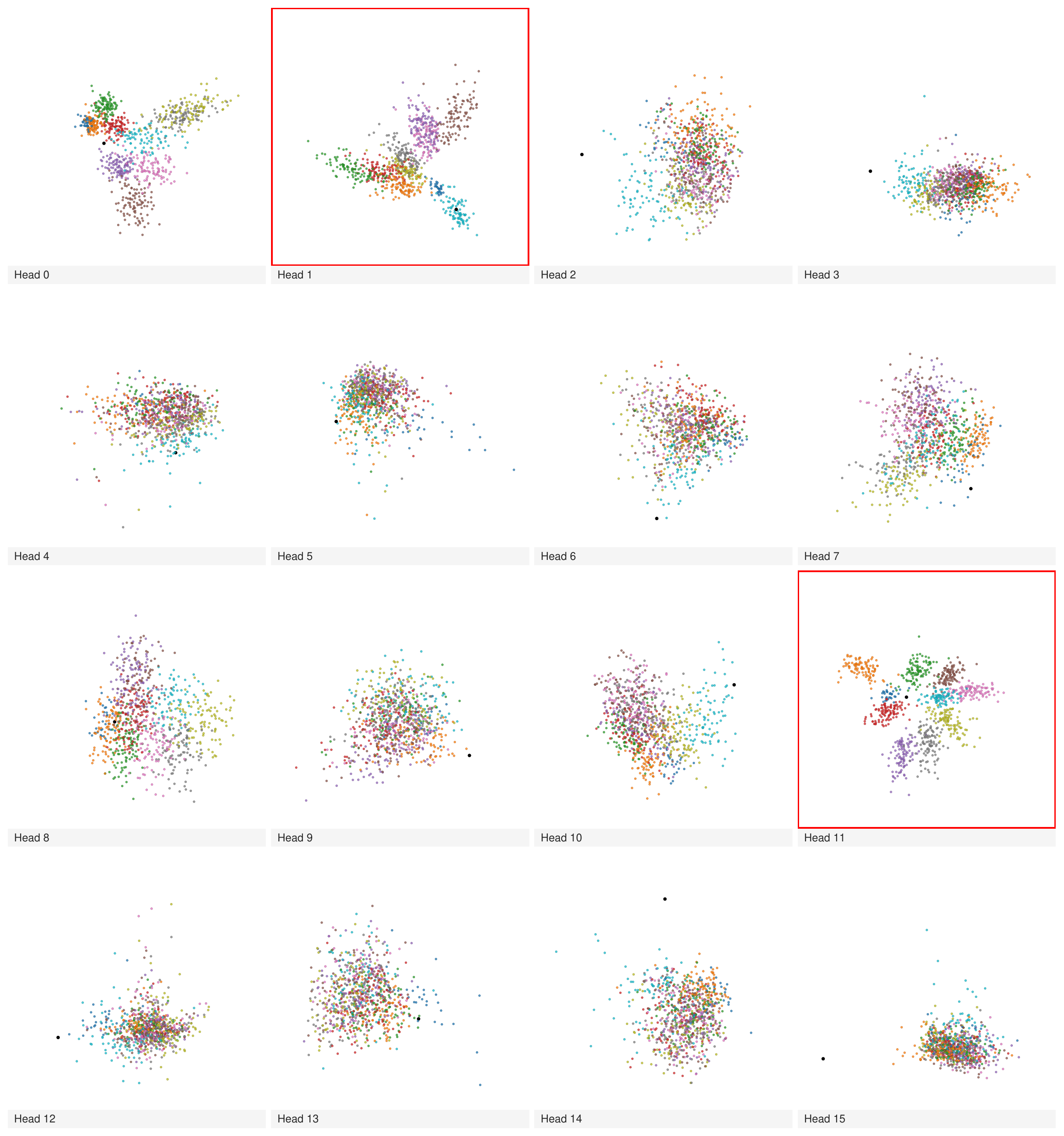}
\caption{Gemma-3-12B-IT.}
\end{subfigure}
\caption{\textbf{Three-dimensional PCA of attention-head outputs across models (continued).} Gemma-3-12B-IT. Colors and red frames follow panel (a).}
\end{figure}

\subsection{Uncertainty estimates}
\label{app:uncertainty}

Tables~\ref{tab:ci-fig3}--\ref{tab:ci-redirection} give 95\% intervals for
every value reported in Figure~\ref{fig:ste-top32-behavior},
Table~\ref{tab:cross-position-patching}, and
Table~\ref{tab:router-redirection-proportions}. Intervals are computed from
response counts aggregated per condition and, for the two tables, per ordered
injection--readout position pair.

\paragraph{Wilson intervals.} For each rate we report the 95\% Wilson score
interval. An injected-run rate pools 100 concepts $\times$ 30 prompts $\times$
10 positions $=30{,}000$ trials per condition and label setting (270{,}000 for
the 90 pairs $i\ne j$ in each setting of Table~\ref{tab:cross-position-patching}
and in each arm of Table~\ref{tab:router-redirection-proportions}). Each bar of
Figure~\ref{fig:ste-top32-behavior} and each entry of
Table~\ref{tab:cross-position-patching} is the unweighted mean over the six
label settings; because every setting contributes the same number of trials,
this mean equals the rate pooled over the settings, and its interval is the
Wilson interval of the pooled counts: 180{,}000 trials for most bars,
1{,}800{,}000 for the two bars that pool all 100 gate--router position pairs,
and 1{,}620{,}000 for Table~\ref{tab:cross-position-patching}.
This gives half-widths of at most $0.3$ percentage points for every
Figure~\ref{fig:ste-top32-behavior} bar other than the unmodified clean run, and
$0.002$ in Table~\ref{tab:router-redirection-proportions}. These intervals
describe sampling over trials, not how much a rate varies across label
settings. The unmodified clean run depends on neither the concept nor the
injection position, so its rate is determined by the test prompts; its
intervals use $n=30$ per setting, 180 in Figure~\ref{fig:ste-top32-behavior}. For the pooled tables we
additionally report a position-level bootstrap.

\paragraph{Position bootstrap.} Table~\ref{tab:cross-position-patching} and
Table~\ref{tab:router-redirection-proportions} pool over the 90 ordered
position pairs. We resample the ten source positions $i$ with replacement, keep
all nine pairs of every sampled position, and recompute the pooled rate
(10{,}000 resamples, percentile intervals). For
Table~\ref{tab:cross-position-patching} and the Mean column of
Table~\ref{tab:router-redirection-proportions}, the six label settings share
each resample of positions. These intervals are deliberately conservative.

\paragraph{Results.} The main comparisons hold under these intervals. In Figure~\ref{fig:ste-top32-behavior}a,b, the gate patch moves each
rate well outside the interval of the unmodified run in all three models. In
Figure~\ref{fig:ste-top32-behavior}d, the \texttt{none} rate with clean gate
heads and injected router heads lies far above the interval of the unmodified
injected run. In Table~\ref{tab:cross-position-patching}, output $j$ exceeds
output $i$ under both intervals in every model. In
Table~\ref{tab:router-redirection-proportions}, ${\to}\,j$ is the most
frequent outcome for all three models when averaged over arms.

\begin{table}[htbp]
  \providecommand{\ciint}[2]{\textcolor{black!55}{[#1,\,#2]}}
  \centering
  \caption{\textbf{95\% intervals for Figure~\ref{fig:ste-top32-behavior}.}
  Plotted rate (\%) and its Wilson interval. Each rate is the mean over the
  six label settings of Table~\ref{tab:task-performance}, which equals the rate
  pooled over them. Unless marked, a rate pools 180{,}000 trials (six settings
  $\times$ 100 concepts $\times$ 30 prompts $\times$ 10 positions).
  $^\ast$~Unmodified clean run: 180 prompts (30 per setting).
  $^\dagger$~1{,}800{,}000 trials (all 100 gate--router position pairs in each
  setting).
  \textbf{Bold}: the gate heads alone are patched.}
  \label{tab:ci-fig3}
  \small
  \setlength{\tabcolsep}{3.5pt}
  \begin{tabular*}{\textwidth}{@{}l@{\extracolsep{\fill}}r@{\extracolsep{0pt}\hspace{2\tabcolsep}}>{\scriptsize}l@{\extracolsep{\fill}}r@{\extracolsep{0pt}\hspace{2\tabcolsep}}>{\scriptsize}l@{\extracolsep{\fill}}r@{\extracolsep{0pt}\hspace{2\tabcolsep}}>{\scriptsize}l@{}}
    \toprule
    & \multicolumn{2}{c}{Qwen3-4B-IT} & \multicolumn{2}{c}{LLaMA-3.1-8B-IT} & \multicolumn{2}{c}{Gemma-3-12B-IT} \\
    \cmidrule(lr){2-3}\cmidrule(lr){4-5}\cmidrule(l){6-7}
    Condition & \multicolumn{1}{c}{\%} & \multicolumn{1}{c}{95\% CI}
              & \multicolumn{1}{c}{\%} & \multicolumn{1}{c}{95\% CI}
              & \multicolumn{1}{c}{\%} & \multicolumn{1}{c}{95\% CI} \\
    \midrule
    \multicolumn{7}{@{}l}{\textbf{(a)} Gate off: \texttt{none} rate} \\
    \quad Injected run
      & 37.7 & \ciint{37.5}{37.9} & 25.3 & \ciint{25.1}{25.5} & 23.8 & \ciint{23.6}{24.0} \\
    \quad Gate patch
      & \textbf{81.5} & \ciint{81.4}{81.7} & \textbf{60.7} & \ciint{60.5}{60.9} & \textbf{80.7} & \ciint{80.5}{80.9} \\
    \quad Target (clean run)$^\ast$
      & 89.4 & \ciint{84.1}{93.1} & 62.8 & \ciint{55.5}{69.5} & 87.8 & \ciint{82.2}{91.8} \\
    \addlinespace[3pt]
    \multicolumn{7}{@{}l}{\textbf{(b)} Gate on: position rate} \\
    \quad Clean run$^\ast$
      & 10.6 & \ciint{6.9}{15.9} & 37.2 & \ciint{30.5}{44.5} & 12.2 & \ciint{8.2}{17.8} \\
    \quad Gate patch
      & \textbf{41.8} & \ciint{41.5}{42.0} & \textbf{69.3} & \ciint{69.1}{69.6} & \textbf{62.6} & \ciint{62.3}{62.8} \\
    \quad Target (injected run)
      & 62.3 & \ciint{62.1}{62.5} & 74.7 & \ciint{74.5}{74.9} & 76.2 & \ciint{76.0}{76.4} \\
    \addlinespace[3pt]
    \multicolumn{7}{@{}l}{\textbf{(c)} Clean run: position rate} \\
    \quad Clean run$^\ast$
      & 10.6 & \ciint{6.9}{15.9} & 37.2 & \ciint{30.5}{44.5} & 12.2 & \ciint{8.2}{17.8} \\
    \quad Gate clean $\times$ router injected
      & 14.1 & \ciint{13.9}{14.2} & 42.7 & \ciint{42.5}{43.0} & 9.8 & \ciint{9.6}{9.9} \\
    \quad Gate injected $\times$ router clean
      & \textbf{36.2} & \ciint{36.0}{36.4} & \textbf{67.7} & \ciint{67.4}{67.9} & \textbf{42.7} & \ciint{42.4}{42.9} \\
    \quad Gate injected $\times$ router injected$^\dagger$
      & 50.3 & \ciint{50.2}{50.3} & 73.7 & \ciint{73.6}{73.7} & 61.7 & \ciint{61.7}{61.8} \\
    \addlinespace[3pt]
    \multicolumn{7}{@{}l}{\textbf{(d)} Injected run: \texttt{none} rate} \\
    \quad Injected run
      & 37.7 & \ciint{37.5}{37.9} & 25.3 & \ciint{25.1}{25.5} & 23.8 & \ciint{23.6}{24.0} \\
    \quad Gate injected $\times$ router clean
      & 49.3 & \ciint{49.0}{49.5} & 31.6 & \ciint{31.4}{31.8} & 38.2 & \ciint{38.0}{38.4} \\
    \quad Gate clean $\times$ router injected$^\dagger$
      & \textbf{76.3} & \ciint{76.3}{76.4} & \textbf{53.9} & \ciint{53.8}{54.0} & \textbf{73.7} & \ciint{73.6}{73.8} \\
    \quad Gate clean $\times$ router clean
      & 83.4 & \ciint{83.2}{83.6} & 61.4 & \ciint{61.2}{61.6} & 79.4 & \ciint{79.2}{79.5} \\
    \bottomrule
  \end{tabular*}
\end{table}

\begin{table}[htbp]
  \providecommand{\ciint}[2]{\textcolor{black!55}{[#1,\,#2]}}
  \centering
  \caption{\textbf{95\% intervals for Table~\ref{tab:cross-position-patching}.}
  Rate (\%) averaged over the six label settings of
  Table~\ref{tab:task-performance}, each pooling the 90 ordered pairs $i\ne j$
  (1{,}620{,}000 trials in total), with two intervals: \emph{Wilson}, a score
  interval over trials, and \emph{Pos.\ boot.}, a bootstrap that resamples the
  ten gate-source positions $i$ together with all of their pairs, shared across
  the six settings. Other is the average rate per position other than $i$
  and $j$, and its intervals are those of the eight-position total divided
  by eight. Output $j$ exceeds output $i$ and Other under
  both intervals in every model.}
  \label{tab:ci-cross-position}
  \small
  \setlength{\tabcolsep}{3pt}
  \begin{tabular*}{\textwidth}{@{}l@{\extracolsep{\fill}}r@{\extracolsep{0pt}\hspace{2\tabcolsep}}>{\scriptsize}c@{\extracolsep{0pt}\hspace{2\tabcolsep}}>{\scriptsize}c@{\extracolsep{\fill}}r@{\extracolsep{0pt}\hspace{2\tabcolsep}}>{\scriptsize}c@{\extracolsep{0pt}\hspace{2\tabcolsep}}>{\scriptsize}c@{\extracolsep{\fill}}r@{\extracolsep{0pt}\hspace{2\tabcolsep}}>{\scriptsize}c@{\extracolsep{0pt}\hspace{2\tabcolsep}}>{\scriptsize}c@{}}
    \toprule
    & \multicolumn{3}{c}{Qwen3-4B-IT} & \multicolumn{3}{c}{LLaMA-3.1-8B-IT} & \multicolumn{3}{c}{Gemma-3-12B-IT} \\
    \cmidrule(lr){2-4}\cmidrule(lr){5-7}\cmidrule(l){8-10}
    Outcome & \multicolumn{1}{c}{\%} & \multicolumn{1}{c}{\footnotesize Wilson} & \multicolumn{1}{c}{\footnotesize Pos.\ boot.}
            & \multicolumn{1}{c}{\%} & \multicolumn{1}{c}{\footnotesize Wilson} & \multicolumn{1}{c}{\footnotesize Pos.\ boot.}
            & \multicolumn{1}{c}{\%} & \multicolumn{1}{c}{\footnotesize Wilson} & \multicolumn{1}{c}{\footnotesize Pos.\ boot.} \\
    \midrule
    Output $i$
      & 2.9 & \ciint{2.9}{2.9} & \ciint{2.2}{3.7}
      & 3.2 & \ciint{3.2}{3.2} & \ciint{1.5}{5.8}
      & 4.9 & \ciint{4.9}{5.0} & \ciint{1.2}{10.3} \\
    Output $j$
      & \textbf{38.0} & \ciint{37.9}{38.1} & \ciint{32.3}{42.6}
      & \textbf{37.8} & \ciint{37.7}{37.9} & \ciint{35.4}{39.4}
      & \textbf{40.1} & \ciint{40.0}{40.1} & \ciint{31.0}{47.4} \\
    Other
      & 1.1 & \ciint{1.1}{1.1} & \ciint{0.9}{1.3}
      & 4.1 & \ciint{4.1}{4.1} & \ciint{3.6}{4.4}
      & 2.0 & \ciint{2.0}{2.0} & \ciint{1.5}{2.3} \\
    \bottomrule
  \end{tabular*}
\end{table}

\begin{table}[htbp]
  \providecommand{\ciint}[2]{\textcolor{black!55}{[#1,\,#2]}}
  \providecommand{\cistack}[3]{%
    \renewcommand{\arraystretch}{1}\begin{tabular}{@{}c@{}}#1\\[-1.5pt]{\scriptsize\ciint{#2}{#3}}\end{tabular}}
  \centering
  \caption{\textbf{95\% position-bootstrap intervals for
  Table~\ref{tab:router-redirection-proportions}.} Each cell gives the
  proportion from Table~\ref{tab:router-redirection-proportions} and, below
  it, an interval from resampling the ten injection positions $i$ with all of
  their readout positions $j\ne i$; the Mean column resamples positions
  jointly across the six arms. Wilson half-widths are at most 0.002 for every
 a score interval over trials, and  arm and at most 0.001 for the Mean.}
  \label{tab:ci-redirection}
  \small
  \setlength{\tabcolsep}{4pt}
  \renewcommand{\arraystretch}{1.15}
  \begin{tabular*}{\textwidth}{@{}l@{\extracolsep{\fill}} cccccc c@{}}
    \toprule
    & \multicolumn{3}{c}{Ordered labels} & \multicolumn{3}{c}{Shuffled labels} & \\
    \cmidrule(lr){2-4}\cmidrule(lr){5-7}
    Outcome & Digits & Letters & Words & Digits & Letters & Words & \textbf{Mean} \\
    \midrule
    \multicolumn{8}{@{}l}{\textit{Qwen3-4B-IT}} \\
    \quad $\to j$
      & \cistack{0.480}{0.361}{0.587} & \cistack{0.679}{0.560}{0.776} & \cistack{0.704}{0.624}{0.769}
      & \cistack{0.582}{0.491}{0.652} & \cistack{0.585}{0.532}{0.631} & \cistack{0.760}{0.734}{0.784}
      & \cistack{\textbf{0.632}}{0.552}{0.695} \\ \addlinespace[2.5pt]
    \quad $\to i$
      & \cistack{0.011}{0.001}{0.027} & \cistack{0.011}{0.005}{0.019} & \cistack{0.029}{0.016}{0.045}
      & \cistack{0.004}{0.001}{0.008} & \cistack{0.018}{0.012}{0.024} & \cistack{0.023}{0.016}{0.030}
      & \cistack{\textbf{0.016}}{0.013}{0.020} \\ \addlinespace[2.5pt]
    \quad Other
      & \cistack{0.004}{0.002}{0.005} & \cistack{0.003}{0.002}{0.004} & \cistack{0.022}{0.014}{0.031}
      & \cistack{0.013}{0.009}{0.017} & \cistack{0.031}{0.025}{0.035} & \cistack{0.065}{0.053}{0.076}
      & \cistack{\textbf{0.023}}{0.019}{0.027} \\ \addlinespace[2.5pt]
    \quad \texttt{none}
      & \cistack{0.505}{0.396}{0.627} & \cistack{0.307}{0.209}{0.427} & \cistack{0.245}{0.176}{0.333}
      & \cistack{0.401}{0.330}{0.494} & \cistack{0.366}{0.315}{0.427} & \cistack{0.152}{0.129}{0.179}
      & \cistack{\textbf{0.329}}{0.263}{0.413} \\
    \midrule
    \multicolumn{8}{@{}l}{\textit{LLaMA-3.1-8B-IT}} \\
    \quad $\to j$
      & \cistack{0.759}{0.738}{0.784} & \cistack{0.530}{0.448}{0.601} & \cistack{0.680}{0.655}{0.704}
      & \cistack{0.829}{0.819}{0.843} & \cistack{0.676}{0.661}{0.693} & \cistack{0.725}{0.713}{0.739}
      & \cistack{\textbf{0.700}}{0.681}{0.717} \\ \addlinespace[2.5pt]
    \quad $\to i$
      & \cistack{0.056}{0.028}{0.085} & \cistack{0.023}{0.012}{0.035} & \cistack{0.059}{0.031}{0.089}
      & \cistack{0.030}{0.013}{0.055} & \cistack{0.023}{0.017}{0.029} & \cistack{0.044}{0.030}{0.060}
      & \cistack{\textbf{0.039}}{0.032}{0.047} \\ \addlinespace[2.5pt]
    \quad Other
      & \cistack{0.078}{0.059}{0.099} & \cistack{0.009}{0.008}{0.011} & \cistack{0.040}{0.025}{0.055}
      & \cistack{0.130}{0.107}{0.149} & \cistack{0.044}{0.037}{0.050} & \cistack{0.189}{0.168}{0.207}
      & \cistack{\textbf{0.082}}{0.071}{0.091} \\ \addlinespace[2.5pt]
    \quad \texttt{none}
      & \cistack{0.107}{0.082}{0.143} & \cistack{0.437}{0.359}{0.526} & \cistack{0.221}{0.188}{0.266}
      & \cistack{0.011}{0.008}{0.014} & \cistack{0.258}{0.241}{0.273} & \cistack{0.042}{0.034}{0.049}
      & \cistack{\textbf{0.179}}{0.158}{0.207} \\
    \midrule
    \multicolumn{8}{@{}l}{\textit{Gemma-3-12B-IT}} \\
    \quad $\to j$
      & \cistack{0.355}{0.264}{0.443} & \cistack{0.524}{0.416}{0.609} & \cistack{0.334}{0.252}{0.403}
      & \cistack{0.322}{0.285}{0.353} & \cistack{0.428}{0.379}{0.472} & \cistack{0.199}{0.162}{0.229}
      & \cistack{\textbf{0.360}}{0.300}{0.408} \\ \addlinespace[2.5pt]
    \quad $\to i$
      & \cistack{0.467}{0.383}{0.548} & \cistack{0.170}{0.120}{0.231} & \cistack{0.394}{0.315}{0.481}
      & \cistack{0.295}{0.241}{0.349} & \cistack{0.151}{0.116}{0.187} & \cistack{0.181}{0.113}{0.269}
      & \cistack{\textbf{0.276}}{0.231}{0.320} \\ \addlinespace[2.5pt]
    \quad Other
      & \cistack{0.009}{0.004}{0.017} & \cistack{0.005}{0.003}{0.008} & \cistack{0.009}{0.005}{0.014}
      & \cistack{0.078}{0.052}{0.113} & \cistack{0.011}{0.008}{0.015} & \cistack{0.219}{0.160}{0.270}
      & \cistack{\textbf{0.055}}{0.042}{0.069} \\ \addlinespace[2.5pt]
    \quad \texttt{none}
      & \cistack{0.169}{0.092}{0.281} & \cistack{0.301}{0.217}{0.412} & \cistack{0.263}{0.194}{0.363}
      & \cistack{0.305}{0.255}{0.354} & \cistack{0.411}{0.338}{0.490} & \cistack{0.401}{0.366}{0.437}
      & \cistack{\textbf{0.308}}{0.250}{0.380} \\
    \bottomrule
  \end{tabular*}
\end{table}

\subsection{Gate and router interventions under each label setting}
\label{app:label-settings}

\begin{figure}[!htbp]
  \centering
  \begin{subfigure}{\linewidth}
    \centering
    \includegraphics[width=\linewidth]{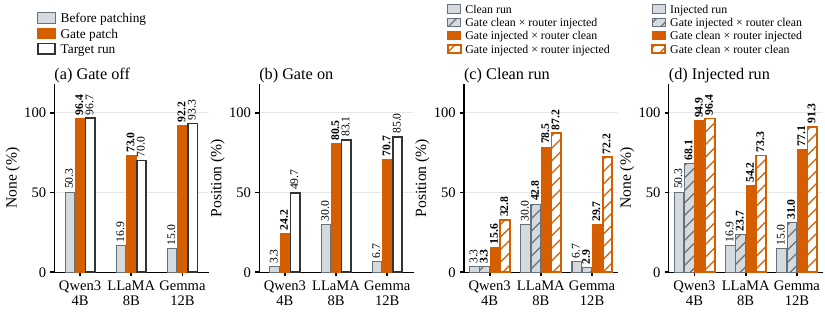}
    \caption{Digits (setting used for head selection).}
  \end{subfigure}
  \par\medskip
  \begin{subfigure}{\linewidth}
    \centering
    \includegraphics[width=\linewidth]{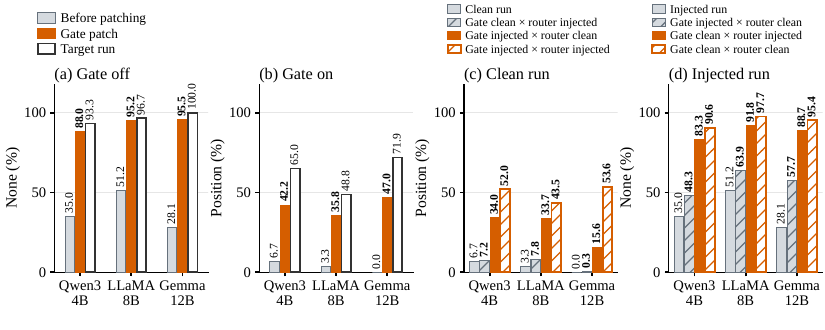}
    \caption{Letters.}
  \end{subfigure}
  \par\medskip
  \begin{subfigure}{\linewidth}
    \centering
    \includegraphics[width=\linewidth]{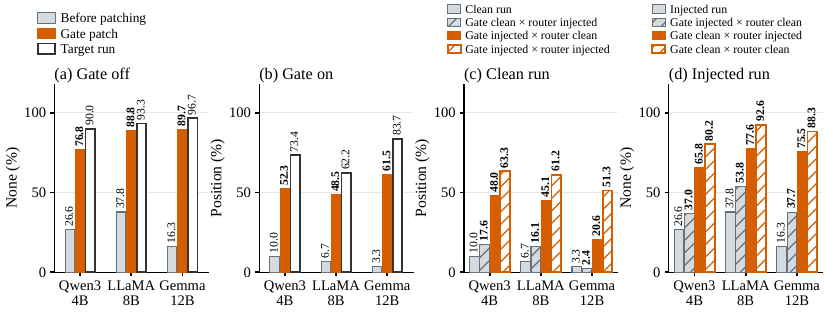}
    \caption{Number words.}
  \end{subfigure}
  \caption{\textbf{Gate and router interventions with ordered labels.}
  Panels as in Figure~\ref{fig:ste-top32-behavior}; gate and router heads
  are the ones selected under ordered digit labels.}
  \label{fig:label-settings-ordered}
\end{figure}

\begin{figure}[!htbp]
  \centering
  \begin{subfigure}{\linewidth}
    \centering
    \includegraphics[width=\linewidth]{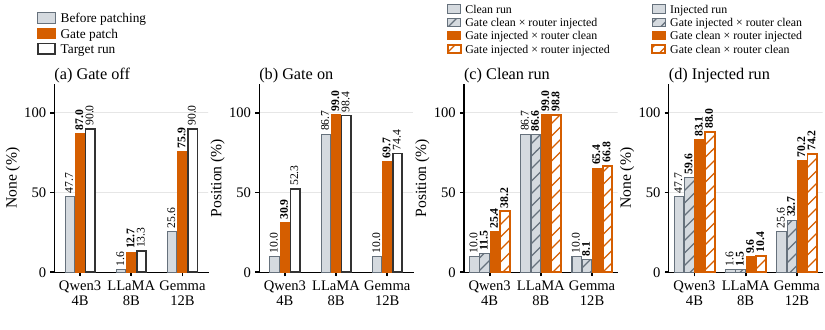}
    \caption{Digits.}
  \end{subfigure}
  \par\medskip
  \begin{subfigure}{\linewidth}
    \centering
    \includegraphics[width=\linewidth]{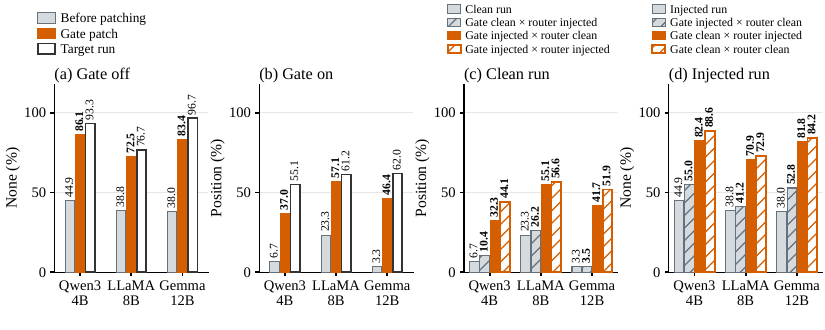}
    \caption{Letters.}
  \end{subfigure}
  \par\medskip
  \begin{subfigure}{\linewidth}
    \centering
    \includegraphics[width=\linewidth]{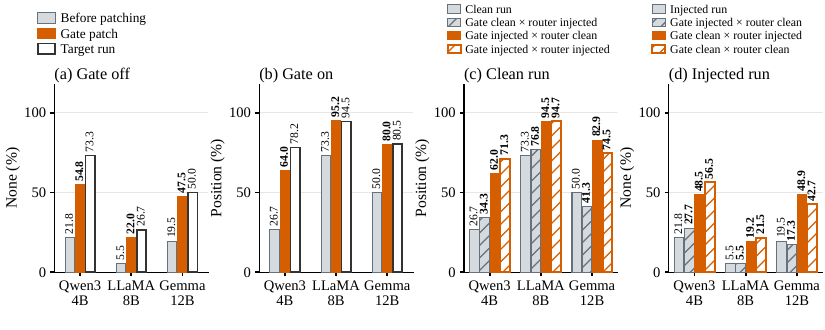}
    \caption{Number words.}
  \end{subfigure}
  \caption{\textbf{Gate and router interventions with shuffled labels.}
  Panels as in Figure~\ref{fig:ste-top32-behavior}; gate and router heads
  are the ones selected under ordered digit labels.}
  \label{fig:label-settings-shuffled}
\end{figure}

Figure~\ref{fig:ste-top32-behavior} and
Table~\ref{tab:cross-position-patching} report the unweighted mean over the
six label settings of Table~\ref{tab:task-performance}.
Figures~\ref{fig:label-settings-ordered} and~\ref{fig:label-settings-shuffled}
show each setting separately, and Table~\ref{tab:cross-position-by-setting}
gives the cross-position patching of each setting.
In every setting the gate and router heads are those selected under ordered
digit labels, without re-selection; ordered digits is the setting the heads
were selected on, and the other five test transfer.
In all eighteen model--setting combinations, the gate patch moves the response
toward the target run (panels a, b), the gate intervention changes the response
more than the router intervention (panels c, d), and output $j$ exceeds
output $i$.

\begin{table}[!htbp]
  \centering
  \caption{\textbf{Cross-position patching under each label setting.}
  Setup as in Table~\ref{tab:cross-position-patching}, whose values are the
  unweighted mean of these six tables. Values are percentages of all trials:
  output $i$, output $j$, Other (the average rate per position other than
  $i$ and $j$, as in Table~\ref{tab:cross-position-patching}), and \texttt{none}.}
  \label{tab:cross-position-by-setting}
  \scriptsize
  \setlength{\tabcolsep}{3pt}
  \begin{subtable}{0.49\linewidth}
    \centering
    \caption{Ordered digits (head selection).}
    \begin{tabular}{@{}lcccc@{}}
      \toprule
      Model & $i$ & $j$ & Other & \texttt{none} \\
      \midrule
      Qwen3-4B-IT & 1.6 & \textbf{28.0} & 0.3 & 68.0 \\
      LLaMA-3.1-8B-IT & 1.4 & \textbf{68.0} & 2.3 & 12.5 \\
      Gemma-3-12B-IT & 4.3 & \textbf{62.8} & 0.5 & 28.5 \\
      \bottomrule
    \end{tabular}
  \end{subtable}
  \hfill
  \begin{subtable}{0.49\linewidth}
    \centering
    \caption{Shuffled digits.}
    \begin{tabular}{@{}lcccc@{}}
      \toprule
      Model & $i$ & $j$ & Other & \texttt{none} \\
      \midrule
      Qwen3-4B-IT & 2.1 & \textbf{27.4} & 1.0 & 62.3 \\
      LLaMA-3.1-8B-IT & 5.4 & \textbf{34.1} & 7.4 & 1.2 \\
      Gemma-3-12B-IT & 5.0 & \textbf{35.2} & 3.3 & 33.3 \\
      \bottomrule
    \end{tabular}
  \end{subtable}
  \par\medskip
  \begin{subtable}{0.49\linewidth}
    \centering
    \caption{Ordered letters.}
    \begin{tabular}{@{}lcccc@{}}
      \toprule
      Model & $i$ & $j$ & Other & \texttt{none} \\
      \midrule
      Qwen3-4B-IT & 2.6 & \textbf{43.1} & 0.8 & 48.3 \\
      LLaMA-3.1-8B-IT & 0.8 & \textbf{35.8} & 0.9 & 56.5 \\
      Gemma-3-12B-IT & 3.0 & \textbf{47.1} & 0.3 & 47.8 \\
      \bottomrule
    \end{tabular}
  \end{subtable}
  \hfill
  \begin{subtable}{0.49\linewidth}
    \centering
    \caption{Shuffled letters.}
    \begin{tabular}{@{}lcccc@{}}
      \toprule
      Model & $i$ & $j$ & Other & \texttt{none} \\
      \midrule
      Qwen3-4B-IT & 2.9 & \textbf{33.3} & 1.0 & 56.1 \\
      LLaMA-3.1-8B-IT & 2.9 & \textbf{21.7} & 4.0 & 43.3 \\
      Gemma-3-12B-IT & 2.4 & \textbf{41.1} & 1.0 & 48.6 \\
      \bottomrule
    \end{tabular}
  \end{subtable}
  \par\medskip
  \begin{subtable}{0.49\linewidth}
    \centering
    \caption{Ordered number words.}
    \begin{tabular}{@{}lcccc@{}}
      \toprule
      Model & $i$ & $j$ & Other & \texttt{none} \\
      \midrule
      Qwen3-4B-IT & 3.8 & \textbf{49.6} & 1.2 & 36.9 \\
      LLaMA-3.1-8B-IT & 1.1 & \textbf{46.6} & 1.7 & 38.5 \\
      Gemma-3-12B-IT & 7.7 & \textbf{39.2} & 0.3 & 50.6 \\
      \bottomrule
    \end{tabular}
  \end{subtable}
  \hfill
  \begin{subtable}{0.49\linewidth}
    \centering
    \caption{Shuffled number words.}
    \begin{tabular}{@{}lcccc@{}}
      \toprule
      Model & $i$ & $j$ & Other & \texttt{none} \\
      \midrule
      Qwen3-4B-IT & 4.3 & \textbf{46.5} & 2.6 & 28.7 \\
      LLaMA-3.1-8B-IT & 7.5 & \textbf{20.7} & 8.3 & 5.3 \\
      Gemma-3-12B-IT & 7.2 & \textbf{14.9} & 6.5 & 25.7 \\
      \bottomrule
    \end{tabular}
  \end{subtable}
\end{table}

\clearpage
\section{Comparison of the key and query terms of the QK decomposition}
\label{app:qk-term-comparison}

We first show why the analysis in Section~\ref{sec:intro-vs-non} studies
the score change $\Delta s$. At the final prompt position, the attention
weights are a softmax over the visible positions,
$a_t=e^{s_t}/\sum_u e^{s_u}$. Writing $s^I=s^0+\Delta s$ gives
\begin{equation}
a^I_t
=\frac{e^{s^0_t}e^{\Delta s_t}}{\sum_u e^{s^0_u}e^{\Delta s_u}}
=\frac{a^0_t\,e^{\Delta s_t}}{\sum_u a^0_u\,e^{\Delta s_u}},
\qquad
\log\frac{a^I_t}{a^0_t}
=\Delta s_t-\log\sum_u a^0_u\,e^{\Delta s_u},
\label{eq:attention-reweighting}
\end{equation}
where the second step divides the numerator and denominator by
$\sum_u e^{s^0_u}$. Positions excluded by the attention mask are the
same in both runs, so the identity is exact. The clean run involves no
injection, so for a given prompt and head, $a^0$ is the same for every
concept. By Equation~\ref{eq:attention-reweighting}, $a^I$ is then a
function of $\Delta s$ alone, and it is unchanged when a constant is
added to every $\Delta s_t$. Differences in attention across concepts
therefore arise only through differences in $\Delta s$ across positions.

Equation~\ref{eq:intro-qk-ov-decomposition} splits the injection-induced change
in the attention score into a key term $\Delta K_tq^I/\sqrt{d_h}$ and a query term
$K_t^0\Delta q/\sqrt{d_h}$. With $K_t$ and $q$ the post-RoPE key at position $t$
and query at the final prompt position, substituting $q^0=q^I-\Delta q$ gives
\begin{equation}
\Delta s_t
=\frac{K_t^Iq^I-K_t^0q^0}{\sqrt{d_h}}
=\frac{(K_t^0+\Delta K_t)q^I-K_t^0(q^I-\Delta q)}{\sqrt{d_h}}
=\frac{\Delta K_tq^I+K_t^0\Delta q}{\sqrt{d_h}}.
\end{equation}
This is an identity rather than an approximation, so
either term can be subtracted from the injected forward state on its own. This
appendix reports the distributional and behavioural consequences of doing so,
which motivate our focus on the key term in Section~\ref{sec:intro-vs-non}.

Holding the injected forward state fixed, we subtract either the query term or
the key term from the final row of the attention scores in the selected gate
heads, and measure $D_{\mathrm{KL}}(p\|p_{-X})$ in nats between the resulting
distribution and the unablated one. The successor-position column first
conditions the attention probabilities on the ten successor positions and
compares the resulting relative distribution; the full-context column uses all
visible positions. The output column re-runs the forward pass under the same
score ablation and takes the argmax over the eleven candidate first tokens
\texttt{0}--\texttt{9} and \texttt{none}, reporting the fraction of trials that
name the injected position correctly. $\mathcal C_{\mathrm{intro}}$ is the 100
validation concepts and $\mathcal C_{\mathrm{nonintro}}$ the 100 low-accuracy
concepts (Section~\ref{sec:intro-vs-non}); both are evaluated on the
30 test clusters, giving 30{,}000 trials per concept set. Native rates therefore
differ from the ordered-digit column of Table~\ref{tab:task-performance}, which
uses the test concepts; KL values are computed per trial and
head and then averaged with equal weight.

\begin{table}[H]
\centering
\captionsetup{justification=raggedright,singlelinecheck=false}
\caption{\textbf{Ablating the key term versus the query term.} Attention-distribution change and reporting accuracy under each ablation, by model and concept set.}
\label{tab:qk-term-ablation}
\small
\setlength{\tabcolsep}{3pt}
\renewcommand{\arraystretch}{1.12}
\begin{tabular*}{\linewidth}{@{\extracolsep{\fill}}@{}llcccc@{}}
\toprule
& & \multicolumn{2}{c}{KL: $-$query / $-$key}
& \multicolumn{2}{c}{Correct-position rate} \\
\cmidrule(lr){3-4}\cmidrule(lr){5-6}
Model & Concept set & Successor & Full context
& Native & $-$query / $-$key \\
\midrule
Qwen3-4B-IT & $\mathcal C_{\mathrm{intro}}$ & $0.0197$ / $0.3108$ & $0.0351$ / $0.1581$ & $48.28\%$ & $43.62\%$ / $32.07\%$ \\
& $\mathcal C_{\mathrm{nonintro}}$ & $0.0025$ / $0.0384$ & $0.0043$ / $0.0300$ & $1.67\%$ & $1.49\%$ / $0.96\%$ \\
\midrule
LLaMA-3.1-8B-IT & $\mathcal C_{\mathrm{intro}}$ & $0.0093$ / $0.0585$ & $0.0134$ / $0.0411$ & $69.34\%$ & $67.27\%$ / $65.38\%$ \\
& $\mathcal C_{\mathrm{nonintro}}$ & $0.0041$ / $0.0322$ & $0.0079$ / $0.0266$ & $20.97\%$ & $19.03\%$ / $18.09\%$ \\
\midrule
Gemma-3-12B-IT & $\mathcal C_{\mathrm{intro}}$ & $0.0519$ / $0.5847$ & $0.1205$ / $0.3979$ & $83.67\%$ & $83.07\%$ / $58.23\%$ \\
& $\mathcal C_{\mathrm{nonintro}}$ & $0.0168$ / $0.1751$ & $0.0277$ / $0.1023$ & $18.30\%$ & $17.38\%$ / $6.47\%$ \\
\bottomrule
\end{tabular*}
\par\vspace{4pt}
\begin{minipage}{\linewidth}\footnotesize \textit{Reading.} Each KL cell gives the divergence caused by removing the query term and by removing the key term, respectively; larger values indicate a larger change in the attention distribution. The last column gives the correct-position rate after each ablation, against the native rate in the preceding column.\end{minipage}
\end{table}

Ablating the key term produces the larger distributional change in every model
and concept set, both among the ten successor positions and over the full
context; the full-context key/query ratio of the KL is $3.08$--$7.06\times$.
The output measure orders the two terms the same way in every row. The key-term
effect exceeds the query-term effect in every row: for
$\mathcal C_{\mathrm{intro}}$, removing the key term lowers the
correct-position rate by $3.96$--$25.44$ percentage points, against
$0.60$--$4.66$ points for the query term. This supports concentrating the
structural analysis on the key term.

A change in distribution is not the same as a change in norm. Over the full
context the two terms are of comparable size (query/key $=0.63$--$1.63$), and
the intro/non-intro ratio of the query term ($1.30$--$2.50\times$) is in fact
larger than that of the key term ($1.13$--$1.45\times$). The term
$K_t^0\Delta q$ is a broad, shallow redistribution spread across the context
(effective number of positions $84$--$93$; enrichment over the candidate region
$1.52\times$) and does not concentrate on particular positions.

Two remarks on scope. The decomposition assigns the second-order cross term
$\Delta K_t\Delta q$ to the key term, so ``key term'' throughout denotes the
cross-term-inclusive quantity; for Gemma-3-12B-IT
$\lVert\Delta q\rVert/\lVert q^0\rVert=0.256$, so this term is not negligible in
that model. The conditional KL over the ten successor positions does not reflect
the total attention mass those positions receive, and the behavioural ablation
acts on the entire score row of the selected heads, so it cannot by itself
establish a causal role for the successor positions specifically. The
$\mathcal C_{\mathrm{nonintro}}$ native rate of $1.67\%$ for Qwen3-4B-IT is subject
to a floor effect and its percentage-point changes should not be read alongside
the corresponding intro row.

Section~\ref{sec:intro-vs-non} factors the key response at the ten successor
positions through the SVD $\Delta K_{\mathrm{next}}=\sum_r\sigma_ru_rv_r^\top$.
With $s=\Delta K_{\mathrm{next}}q^I/\sqrt{d_h}$ the full key response and
$U_{i1}$ the entry of $u_1$ at the target position $t_i$, we define
\begin{equation}
s=\sum_r\frac{\sigma_r(v_r^\top q^I)}{\sqrt{d_h}}u_r,
\qquad
R_1=\frac{\sigma_1\,|v_1^\top q^I|}{\sqrt{d_h}},
\qquad
s_i^{(1)}=\frac{\sigma_1\,(v_1^\top q^I)}{\sqrt{d_h}}\,U_{i1},
\qquad
\bar U_{i1}=\frac{\mathbb E\big[s_i^{(1)}\big]}{\mathbb E\big[R_1\big]}.
\label{eq:intro-first-mode}
\end{equation}
$R_1$ is the magnitude of the first-mode response across the ten successor
positions, and $\bar U_{i1}$ is the share of it that lands on $t_i$ with a
consistent sign. The first mode accounts for all but $3\%$ of the target
response $\bar s_i=\mathbb E[s_i]$.
Table~\ref{tab:intro-vs-non-qk-full} gives the group statistics behind the
QK columns of Table~\ref{tab:intro-vs-non-qk}.

\begin{table}[H]
\centering
\captionsetup{justification=raggedright,singlelinecheck=false}
\caption{\textbf{QK responses of introspective and non-introspective concepts.} Group statistics and between-set ratios in gate heads; the ratio rows extend the QK columns of Table~\ref{tab:intro-vs-non-qk}.}
\label{tab:intro-vs-non-qk-full}
\small
\setlength{\tabcolsep}{3pt}
\renewcommand{\arraystretch}{1.12}
\begin{tabular*}{\linewidth}{@{\extracolsep{\fill}}@{}lrrrrrr@{}}
\toprule
Concept set & \shortstack{Query norm\\$\|q^I\|$}
& \shortstack{Magnitude\\$\sigma_1$}
& \shortstack{Readout\\$|v_1^\top q^I|$}
& \shortstack{First mode\\$R_1$}
& \shortstack{Target loading\\$\bar U_{i1}$}
& \shortstack{Target response\\$\bar s_i$} \\
\midrule
\multicolumn{7}{@{}l}{\textbf{(a) Qwen3-4B-IT}} \\
$\mathcal C_{\mathrm{intro}}$ & $14.2870$ & $9.7021$ & $2.0431$ & $1.8697$ & $0.8847$ & $1.7000$ \\
$\mathcal C_{\mathrm{nonintro}}$ & $14.2356$ & $7.9384$ & $1.1333$ & $0.8099$ & $0.3832$ & $0.3203$ \\
\textit{Intro / Non-intro} & $\mathbf{1.00}\times$ & $\mathbf{1.22}\times$ & $\mathbf{1.80}\times$ & $\mathbf{2.31}\times$ & $\mathbf{2.31}\times$ & $\mathbf{5.31}\times$ \\
\midrule
\multicolumn{7}{@{}l}{\textbf{(b) LLaMA-3.1-8B-IT}} \\
$\mathcal C_{\mathrm{intro}}$ & $12.5969$ & $9.4412$ & $1.1510$ & $0.9493$ & $0.6945$ & $0.6774$ \\
$\mathcal C_{\mathrm{nonintro}}$ & $12.6249$ & $8.7674$ & $0.9611$ & $0.7252$ & $0.5165$ & $0.3851$ \\
\textit{Intro / Non-intro} & $\mathbf{1.00}\times$ & $\mathbf{1.08}\times$ & $\mathbf{1.20}\times$ & $\mathbf{1.31}\times$ & $\mathbf{1.34}\times$ & $\mathbf{1.76}\times$ \\
\midrule
\multicolumn{7}{@{}l}{\textbf{(c) Gemma-3-12B-IT}} \\
$\mathcal C_{\mathrm{intro}}$ & $18.5425$ & $19.7339$ & $1.9243$ & $2.4228$ & $0.8841$ & $2.1398$ \\
$\mathcal C_{\mathrm{nonintro}}$ & $18.7912$ & $17.3643$ & $1.2386$ & $1.3325$ & $0.6950$ & $0.9144$ \\
\textit{Intro / Non-intro} & $\mathbf{0.99}\times$ & $\mathbf{1.14}\times$ & $\mathbf{1.55}\times$ & $\mathbf{1.82}\times$ & $\mathbf{1.27}\times$ & $\mathbf{2.34}\times$ \\
\bottomrule
\end{tabular*}
\par\vspace{4pt}
\begin{minipage}{\linewidth}\footnotesize \textit{Definitions.} $\mathcal C_{\mathrm{intro}}$ and $\mathcal C_{\mathrm{nonintro}}$ are the introspective and non-introspective concept sets; all quantities follow Equation~\ref{eq:intro-first-mode}. Entries are group means, except $\bar U_{i1}$, which is a ratio of group means and mixes concentration on the target position with sign consistency across trials (even spread gives $0.32$). \textit{Ratios.} Intro / Non-intro divides the displayed statistics, rounded to two decimals. Because these are ratios of means, columns do not compose exactly; $R_1$ and $\bar U_{i1}$ compose to $\bar s_i$ only up to the residual modes. Ratios are descriptive, not significance tests.\end{minipage}
\end{table}

\clearpage
\section{Output decomposition and modal analysis of the OV circuit}
\label{app:ov-output-decomposition}

This appendix gives details of the OV analysis in Section~\ref{sec:intro-vs-non}. Superscripts $0$ and $I$ denote the unperturbed and injected states, respectively, with $\Delta V=V^I-V^0$ and $\Delta a=a^I-a^0$. We omit sample, layer, and attention-head indices below.

This analysis uses the same two concept sets as the QK analysis in the main text: $\mathcal C_{\mathrm{intro}}$ and $\mathcal C_{\mathrm{nonintro}}$. Each contains 100 concepts. These names denote concept sets partitioned by reporting performance after injection; they do not imply that every injection within a set succeeds or fails. Both concept sets are evaluated on 30 held-out test clusters, with each concept injected at all ten candidate positions in every cluster (300 injections per concept). We analyze the 32 gate heads selected under the gate-on condition; here, gate-on specifies only the head set. Each metric is first averaged over injection positions, candidate-token clusters, and gate heads within each concept, and then averaged within each concept set.

Let $\Delta o=o^I-o^0$ denote the injection-induced change in the attention-head output at the final prompt position $T$. Since $o=W_OV^\top a$, substituting $a^0=a^I-\Delta a$ gives the decomposition of Equation~\ref{eq:intro-qk-ov-decomposition}:
\begin{equation}
\begin{aligned}
\Delta o
&=W_O\big[(V^I)^\top a^I-(V^0)^\top a^0\big]
=W_O\big[(V^0+\Delta V)^\top a^I-(V^0)^\top(a^I-\Delta a)\big]\\
&=W_O(\Delta V)^\top a^I+W_O(V^0)^\top\Delta a .
\end{aligned}
\end{equation}

\clearpage
\subsection{Detailed OV statistics}
\label{app:ov-detailed}

Section~\ref{sec:intro-vs-non} decomposes the $\Delta V$ term of $\Delta o$ through the SVD of $M=W_O(\Delta V)^\top=\sum_k\mu_ky_kx_k^\top$, computed separately for each trial and gate head over all $n$ positions of the context, so that $Ma^I=\sum_k\mu_k(x_k^\top a^I)y_k$ (Section~\ref{sec:intro-vs-non}). Table~\ref{tab:ov-output-stats} lists the attention and output statistics. Besides the norms, it reports two energy fractions of the five leading modes: the matrix energy $\eta_5=\sum_{k\le5}\mu_k^2/\sum_k\mu_k^2$ and the output energy $\rho_5=\sum_{k\le5}\mu_k^2(x_k^\top a^I)^2/\|Ma^I\|^2$. Because the output directions $y_k$ are orthonormal, $\rho_5$ is the fraction of the squared output norm carried by these modes. Table~\ref{tab:ov-mode-stats} lists, for each of the five leading modes, the size $\mu_k$ and the alignment $|\cos(a^I,x_k)|=|x_k^\top a^I|/\|a^I\|$; we take the absolute value because the sign of a singular vector is arbitrary.

\begin{table}[H]
\centering
\captionsetup{justification=raggedright,singlelinecheck=false}
\caption{\textbf{Attention and output statistics of the $\Delta V$ term in gate heads.} Group means over the full context; energies are percentages.}
\label{tab:ov-output-stats}
\small
\setlength{\tabcolsep}{3pt}
\renewcommand{\arraystretch}{1.12}
\begin{tabular*}{\linewidth}{@{\extracolsep{\fill}}@{}lrrrrrrr@{}}
\toprule
& \multicolumn{2}{c}{Attention} & \multicolumn{2}{c}{Change matrix $M$} & \multicolumn{3}{c}{Output $Ma^I$} \\
\cmidrule(lr){2-3}\cmidrule(lr){4-5}\cmidrule(l){6-8}
Concept set & $\|a^I\|$ & $\|\Delta a\|$ & $\|M\|_F$ & $\eta_5$ & $\|Ma^I\|$ & Top-5 norm & $\rho_5$ \\
\midrule
\multicolumn{8}{@{}l}{\textbf{(a) Qwen3-4B-IT}} \\
$\mathcal C_{\mathrm{intro}}$ & 0.367 & 0.085 & 19.3 & 80.1 & 0.912 & 0.841 & 68.7 \\
$\mathcal C_{\mathrm{nonintro}}$ & 0.360 & 0.036 & 16.6 & 82.0 & 0.313 & 0.236 & 49.3 \\
\textit{Intro / Non-intro} & $1.02\times$ & $2.40\times$ & $1.16\times$ & $0.98\times$ & $2.92\times$ & $3.56\times$ & $1.39\times$ \\
\midrule
\multicolumn{8}{@{}l}{\textbf{(b) LLaMA-3.1-8B-IT}} \\
$\mathcal C_{\mathrm{intro}}$ & 0.367 & 0.049 & 6.16 & 78.2 & 0.127 & 0.102 & 53.3 \\
$\mathcal C_{\mathrm{nonintro}}$ & 0.361 & 0.036 & 5.80 & 79.5 & 0.085 & 0.061 & 47.4 \\
\textit{Intro / Non-intro} & $1.02\times$ & $1.37\times$ & $1.06\times$ & $0.98\times$ & $1.49\times$ & $1.66\times$ & $1.13\times$ \\
\midrule
\multicolumn{8}{@{}l}{\textbf{(c) Gemma-3-12B-IT}} \\
$\mathcal C_{\mathrm{intro}}$ & 0.457 & 0.162 & 25.9 & 76.7 & 1.969 & 1.852 & 81.5 \\
$\mathcal C_{\mathrm{nonintro}}$ & 0.430 & 0.073 & 22.9 & 79.1 & 0.770 & 0.628 & 58.2 \\
\textit{Intro / Non-intro} & $1.06\times$ & $2.23\times$ & $1.13\times$ & $0.97\times$ & $2.56\times$ & $2.95\times$ & $1.40\times$ \\
\bottomrule
\end{tabular*}
\end{table}

\begin{table}[H]
\centering
\captionsetup{justification=raggedright,singlelinecheck=false}
\caption{\textbf{Size and alignment of the five leading modes of $M$ in gate heads.} Group means over the full context.}
\label{tab:ov-mode-stats}
\small
\setlength{\tabcolsep}{2.5pt}
\renewcommand{\arraystretch}{1.12}
\begin{tabular*}{\linewidth}{@{\extracolsep{\fill}}@{}lrrrrrrrrrr@{}}
\toprule
& \multicolumn{5}{c}{Size $\mu_k$} & \multicolumn{5}{c}{Alignment $|\cos(a^I,x_k)|$} \\
\cmidrule(lr){2-6}\cmidrule(l){7-11}
Concept set & $k=1$ & 2 & 3 & 4 & 5 & $k=1$ & 2 & 3 & 4 & 5 \\
\midrule
\multicolumn{11}{@{}l}{\textbf{(a) Qwen3-4B-IT}} \\
$\mathcal C_{\mathrm{intro}}$ & 11.3 & 8.24 & 6.45 & 5.33 & 4.48 & 0.149 & 0.126 & 0.104 & 0.097 & 0.087 \\
$\mathcal C_{\mathrm{nonintro}}$ & 10.1 & 7.05 & 5.47 & 4.43 & 3.68 & 0.039 & 0.059 & 0.059 & 0.061 & 0.058 \\
\textit{Intro / Non-intro} & $1.12\times$ & $1.17\times$ & $1.18\times$ & $1.20\times$ & $1.22\times$ & $3.85\times$ & $2.16\times$ & $1.76\times$ & $1.58\times$ & $1.49\times$ \\
\midrule
\multicolumn{11}{@{}l}{\textbf{(b) LLaMA-3.1-8B-IT}} \\
$\mathcal C_{\mathrm{intro}}$ & 3.54 & 2.60 & 2.02 & 1.65 & 1.41 & 0.051 & 0.052 & 0.054 & 0.059 & 0.052 \\
$\mathcal C_{\mathrm{nonintro}}$ & 3.41 & 2.46 & 1.89 & 1.53 & 1.30 & 0.035 & 0.037 & 0.042 & 0.050 & 0.046 \\
\textit{Intro / Non-intro} & $1.04\times$ & $1.05\times$ & $1.07\times$ & $1.08\times$ & $1.08\times$ & $1.48\times$ & $1.40\times$ & $1.28\times$ & $1.20\times$ & $1.13\times$ \\
\midrule
\multicolumn{11}{@{}l}{\textbf{(c) Gemma-3-12B-IT}} \\
$\mathcal C_{\mathrm{intro}}$ & 14.5 & 10.8 & 8.60 & 7.18 & 6.13 & 0.222 & 0.165 & 0.133 & 0.111 & 0.101 \\
$\mathcal C_{\mathrm{nonintro}}$ & 13.3 & 9.59 & 7.50 & 6.14 & 5.19 & 0.076 & 0.085 & 0.079 & 0.075 & 0.073 \\
\textit{Intro / Non-intro} & $1.09\times$ & $1.13\times$ & $1.15\times$ & $1.17\times$ & $1.18\times$ & $2.90\times$ & $1.93\times$ & $1.68\times$ & $1.49\times$ & $1.37\times$ \\
\bottomrule
\end{tabular*}
\end{table}

In all three models, the two groups have nearly the same attention norm ($\le1.06\times$), change-matrix norm ($\le1.16\times$), and matrix energy in the five leading modes. The sizes $\mu_k$ differ by at most $1.22\times$ at every $k$, whereas the alignment is higher for $\mathcal C_{\mathrm{intro}}$ at every $k$, most strongly at $k=1$ ($1.48$--$3.85\times$). As a result, the five leading modes carry a larger share of the output for $\mathcal C_{\mathrm{intro}}$ ($\rho_5$ of $53$--$82\%$ against $47$--$58\%$), and the output norm is $1.49$--$2.92\times$ larger. For every quantity in Tables~\ref{tab:ov-output-stats} and~\ref{tab:ov-mode-stats}, the 95\% concept-level bootstrap interval of the between-group difference excludes zero (5,000 resamples, conditional on the 30 evaluation clusters, without multiple-comparison correction). Ratios of group means are descriptive and do not compose across columns.

\subsection{Causal ablation of the leading OV modes}
\label{app:ov-causal-modes}

We test whether the leading modes support correct-position reports by subtracting their output contributions in the injected run. For each paired clean and injected example and each gate head, we form $M=W_O(\Delta V)^\top$ over all $n$ context positions and order its modes by decreasing singular value $\mu_k$. The leading-five contribution and its complement are
\begin{equation}
 c_{\le5}=\sum_{k=1}^{5}\mu_k(x_k^\top a^I)y_k,
 \qquad c_{>5}=Ma^I-c_{\le5}.
 \label{eq:ov-top5-ablation}
\end{equation}
We separately subtract $c_{\le5}$, $c_{>5}$, or the entire content-change term $Ma^I$ from the attention output projection at the final prompt position, before any attention-output normalization. The intervention is applied jointly to the 32 gate heads. These vectors are fixed from the unablated paired runs; downstream computation is left free. Thus, the modes are defined from the full context, not only the ten candidate injection positions.

Both blocks of Table~\ref{tab:ov-ablation} use all 30 evaluation clusters, with 100 concepts per set and ten injection positions per concept and cluster: 30,000 paired trials per set, or 60,000 per model. The left block removes the two terms of Equation~\ref{eq:intro-qk-ov-decomposition} as full vectors. The right block removes modes of $M$. Correct-position rate is the fraction of trials in which the highest-scoring candidate among the ten positions and \texttt{none} is the injected position; a \texttt{none} response counts as incorrect. Table~\ref{tab:ov-ablation-probability} additionally reports the mean probability of the correct position token under the full-vocabulary softmax, without renormalizing over these eleven candidates.

% Sources: results/ov_causal_ablation_20260924/accelerated/summary_all_models.csv (terms);
% results/ov_top5_causal_full_20260925/summary_all_models.csv (modes and probabilities).

\begin{table}[H]
\centering
\captionsetup{justification=raggedright,singlelinecheck=false}
\caption{\textbf{Correct-position rate (\%) after removing parts of the gate-head output in the injected run.} $-\Delta a$ and $-\Delta V$ remove the corresponding terms of Equation~\ref{eq:intro-qk-ov-decomposition}; $-$top 5 and $-$others remove the five leading modes of $M$ or all remaining modes.}
\label{tab:ov-ablation}
\small
\setlength{\tabcolsep}{2.5pt}
\renewcommand{\arraystretch}{1.12}
\begin{tabular*}{\linewidth}{@{\extracolsep{\fill}}@{}llrrrrrrrr@{}}
\toprule
& & \multicolumn{4}{c}{Terms, 30 clusters} & \multicolumn{4}{c}{Modes of $M$, 30 clusters} \\
\cmidrule(lr){3-6}\cmidrule(l){7-10}
Model & Set & Injected & $-\Delta a$ & $-\Delta V$ & $-$both & Injected & $-$top 5 & $-$others & $-\Delta V$ \\
\midrule
Qwen3-4B-IT & $\mathcal C_{\mathrm{intro}}$ & 48.3 & 38.8 & 4.4 & 1.4 & 48.3 & 5.1 & 44.6 & 4.4 \\
 & $\mathcal C_{\mathrm{nonintro}}$ & 1.7 & 1.3 & 0.7 & 0.7 & 1.7 & 0.7 & 1.5 & 0.7 \\
\midrule
LLaMA-3.1-8B-IT & $\mathcal C_{\mathrm{intro}}$ & 69.3 & 67.8 & 54.3 & 49.8 & 69.3 & 57.4 & 67.1 & 54.3 \\
 & $\mathcal C_{\mathrm{nonintro}}$ & 21.0 & 19.3 & 13.8 & 11.9 & 21.0 & 15.3 & 19.0 & 13.8 \\
\midrule
Gemma-3-12B-IT & $\mathcal C_{\mathrm{intro}}$ & 83.7 & 74.1 & 13.9 & 2.6 & 83.7 & 13.5 & 81.2 & 13.9 \\
 & $\mathcal C_{\mathrm{nonintro}}$ & 18.3 & 11.0 & 1.5 & 0.4 & 18.3 & 1.8 & 15.4 & 1.5 \\
\bottomrule
\end{tabular*}
\end{table}

\begin{table}[H]
\centering
\captionsetup{justification=raggedright,singlelinecheck=false}
\caption{\textbf{Mean correct-position token probability (\%) after OV mode ablation.} Full-context modes, all 30 clusters, and 30,000 paired trials per concept set and model. Probabilities use the full vocabulary.}
\label{tab:ov-ablation-probability}
\small
\setlength{\tabcolsep}{4pt}
\renewcommand{\arraystretch}{1.12}
\begin{tabular*}{\linewidth}{@{\extracolsep{\fill}}@{}llrrrr@{}}
\toprule
Model & Set & Injected & $-$top 5 & $-$others & $-\Delta V$ \\
\midrule
Qwen3-4B-IT & $\mathcal C_{\mathrm{intro}}$ & 47.99 & 5.12 & 44.39 & 4.40 \\
 & $\mathcal C_{\mathrm{nonintro}}$ & 1.66 & 0.67 & 1.47 & 0.65 \\
\midrule
LLaMA-3.1-8B-IT & $\mathcal C_{\mathrm{intro}}$ & 40.22 & 33.69 & 39.95 & 32.93 \\
 & $\mathcal C_{\mathrm{nonintro}}$ & 13.37 & 11.80 & 12.96 & 11.32 \\
\midrule
Gemma-3-12B-IT & $\mathcal C_{\mathrm{intro}}$ & 83.12 & 13.53 & 80.63 & 13.91 \\
 & $\mathcal C_{\mathrm{nonintro}}$ & 18.04 & 1.83 & 15.15 & 1.43 \\
\bottomrule
\end{tabular*}
\end{table}

For $\mathcal C_{\mathrm{intro}}$, removing the leading five modes lowers correct-token probability from 47.99\% to 5.12\% in Qwen3-4B-IT and from 83.12\% to 13.53\% in Gemma-3-12B-IT. The reduction is smaller in LLaMA-3.1-8B-IT, from 40.22\% to 33.69\%. In each model, removing the remaining modes produces a much smaller probability decrease, whereas removing the leading five yields a result close to removing the entire $\Delta V$ term. Correct-position rates follow the same pattern. These interventions support a causal role for the leading modes in correct-position reporting, with substantial differences in effect size across models; their effects need not add because downstream computation is nonlinear.

The accuracy decreases also involve fewer numeric reports. For $\mathcal C_{\mathrm{intro}}$, the rate of selecting any position rather than \texttt{none} falls from 49.66\% to 6.86\% in Qwen3-4B-IT, from 81.04\% to 67.58\% in LLaMA-3.1-8B-IT, and from 84.46\% to 14.16\% in Gemma-3-12B-IT after leading-five removal. The intervention therefore affects whether a position is reported as well as whether it is correct. The native correct-position rate of $\mathcal C_{\mathrm{nonintro}}$ in Qwen3-4B-IT is near floor, limiting interpretation of its small absolute changes.

\paragraph{Varying the number of removed modes.}
Table~\ref{tab:ov-rank-sweep} extends the same intervention to $c_{\le k}=\sum_{j=1}^{k}\mu_j(x_j^\top a^I)y_j$ for $k=1,2,5,10$. All settings use the same 30 clusters, concepts, positions, and gate heads; the new runs reproduce the previous unablated predictions and correct-token probabilities exactly. Increasing $k$ from five to ten changes mean correct-token probability by less than one percentage point in either concept set for every model, although the effect is not strictly monotonic in Gemma-3-12B-IT.

% Source: results/ov_ksweep_20260925/comparison_all_models.csv.
\begin{table}[H]
\centering
\captionsetup{justification=raggedright,singlelinecheck=false}
\caption{\textbf{OV ablation as a function of the number of removed modes.} Modes of the full-context matrix $M$ are ordered by singular value. Each entry uses all 30 clusters and 30,000 trials per concept set and model. All values are percentages; probabilities use the full-vocabulary softmax. $-\Delta V$ removes the entire content-change term.}
\label{tab:ov-rank-sweep}
\small
\setlength{\tabcolsep}{3pt}
\renewcommand{\arraystretch}{1.12}
\begin{tabular*}{\linewidth}{@{\extracolsep{\fill}}@{}llrrrrrr@{}}
\toprule
Model & Set & Injected & $-$top 1 & $-$top 2 & $-$top 5 & $-$top 10 & $-\Delta V$ \\
\midrule
\multicolumn{8}{@{}l}{\textbf{(a) Mean correct-position token probability}} \\
Qwen3-4B-IT & $\mathcal C_{\mathrm{intro}}$ & 47.99 & 24.89 & 10.28 & 5.12 & 4.42 & 4.40 \\
 & $\mathcal C_{\mathrm{nonintro}}$ & 1.66 & 1.06 & 0.78 & 0.67 & 0.64 & 0.65 \\
\addlinespace[2pt]
LLaMA-3.1-8B-IT & $\mathcal C_{\mathrm{intro}}$ & 40.22 & 38.57 & 36.64 & 33.69 & 33.16 & 32.93 \\
 & $\mathcal C_{\mathrm{nonintro}}$ & 13.37 & 13.16 & 12.82 & 11.80 & 11.46 & 11.32 \\
\addlinespace[2pt]
Gemma-3-12B-IT & $\mathcal C_{\mathrm{intro}}$ & 83.12 & 54.86 & 27.82 & 13.53 & 14.24 & 13.91 \\
 & $\mathcal C_{\mathrm{nonintro}}$ & 18.04 & 10.94 & 5.53 & 1.83 & 1.51 & 1.43 \\
\midrule
\multicolumn{8}{@{}l}{\textbf{(b) Correct-position rate}} \\
Qwen3-4B-IT & $\mathcal C_{\mathrm{intro}}$ & 48.28 & 25.07 & 10.28 & 5.15 & 4.42 & 4.40 \\
 & $\mathcal C_{\mathrm{nonintro}}$ & 1.67 & 1.06 & 0.81 & 0.68 & 0.64 & 0.65 \\
\addlinespace[2pt]
LLaMA-3.1-8B-IT & $\mathcal C_{\mathrm{intro}}$ & 69.34 & 66.72 & 64.05 & 57.42 & 55.37 & 54.27 \\
 & $\mathcal C_{\mathrm{nonintro}}$ & 20.97 & 19.97 & 18.61 & 15.26 & 13.98 & 13.81 \\
\addlinespace[2pt]
Gemma-3-12B-IT & $\mathcal C_{\mathrm{intro}}$ & 83.67 & 55.82 & 28.19 & 13.47 & 14.17 & 13.88 \\
 & $\mathcal C_{\mathrm{nonintro}}$ & 18.30 & 11.11 & 5.55 & 1.83 & 1.52 & 1.45 \\
\bottomrule
\end{tabular*}
\end{table}

\end{document}